\documentclass[lettersize,journal]{IEEEtran}
\usepackage{amsmath,amsfonts}
\usepackage{algorithmic}
\usepackage{algorithm}
\usepackage{array}
\usepackage[caption=false,font=normalsize,labelfont=sf,textfont=sf]{subfig}
\usepackage{textcomp}
\usepackage{stfloats}
\usepackage{url}
\usepackage{verbatim}
\usepackage{graphicx}
\usepackage{cite}
\usepackage{xcolor}
\newcommand{\revise}[1]{\textcolor{black}{#1}}

\newcommand{\mypara}[1]{\noindent\textbf{#1.}~}
\usepackage{enumitem}
\usepackage{multirow}
\usepackage{makecell}
\usepackage{graphicx}
\usepackage{booktabs}
\usepackage[citecolor=blue, colorlinks]{hyperref}
\usepackage{wrapfig}
\usepackage{cuted}
\usepackage{capt-of}
\usepackage{tablefootnote}

\begin{document}

\title{G-Skin: Learning to Bind 3D Gaussians with Generative Visual Priors}

\author{Yuxin Yao, Kendong Liu, Shiqi Zhou, Jiazhi Xia, Junhui Hou,~\IEEEmembership{Senior Member,~IEEE,}
\IEEEcompsocitemizethanks{This work was supported in part by the National Natural Science Foundation of China under Grants 62422118, U23A20313, and 62372471, and in part by the Hong Kong Research Grants Council under Grants 11220426, 11219324, N\_CityU1114/25, and RFS2627-1S02. (\textit{Corresponding author: Junhui Hou})

\IEEEcompsocthanksitem Y. Yao, K. Liu, and J. Hou are with the Department of Computer Science, City University of Hong Kong, Hong Kong SAR, China (email: yuxinyao@cityu.edu.hk; kdliu2-c@my.cityu.edu.hk; jh.hou@cityu.edu.hk). 
\IEEEcompsocthanksitem S. Zhou is with Central Media Technology Institute, Huawei, China (email: zhoushiqi3@huawei.com).
\IEEEcompsocthanksitem J. Xia is with the School of Computer Science and Engineering, Central South University, Changsha, China (email:xiajiazhi@csu.edu.cn).
}}

\markboth{G-Skin: Learning to Bind 3D Gaussians with Generative Visual Priors}%
{Yao \MakeLowercase{\textit{et al.}}: G-Skin}

\IEEEaftertitletext{%
  \noindent
  \begin{minipage}{\textwidth}
    \centering
    \includegraphics[width=0.92\linewidth]{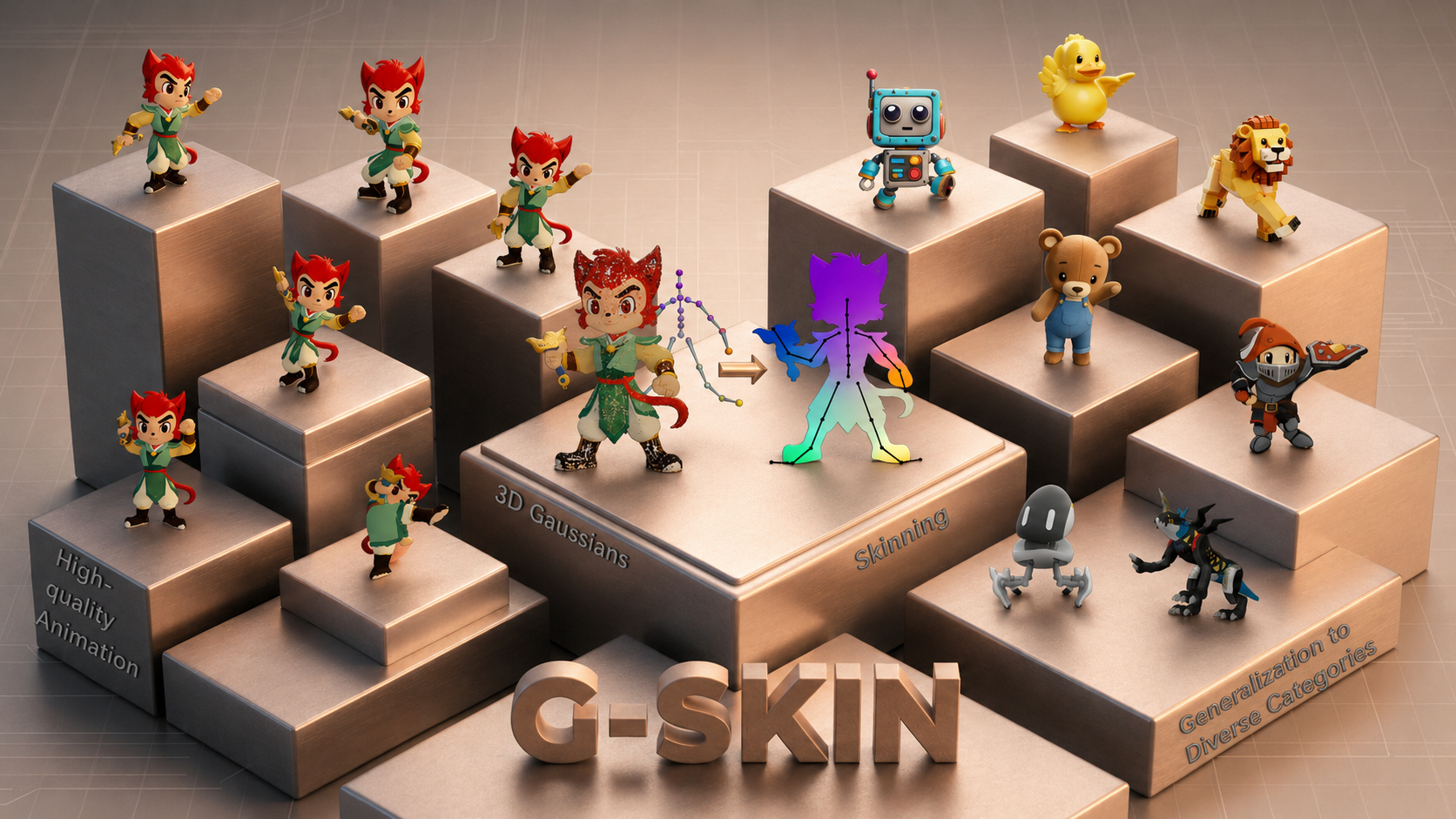}\par
    \captionof{figure}{G-Skin optimizes skinning weights for 3D Gaussian primitives and a given skeleton guided by a generative visual prior, while incorporating local geometric constraints to ensure both realistic animation and minimized rendering artifacts. Refer to the \textcolor{magenta}{\textbf{video demo}} on the project page for more results.}
    \label{fig:teaser}
  \end{minipage}\par
}

\maketitle

\begin{abstract}
3D Gaussian Splatting has achieved remarkable success in photorealistic and efficient rendering, leading to a rapid increase in 3D assets represented by 3D Gaussian primitives. 
\revise{Automatically estimating skinning weights for these assets given skeletons with arbitrary topologies is highly desirable. }
However, training a feed-forward skinning framework is infeasible due to the lack of high-quality 3D Gaussian rigging datasets. An alternative solution is to transfer mesh-based techniques to 3D Gaussian-based representation, but 3D Gaussian primitives are not restricted to the surface and lack explicit topological connectivity. Moreover, this kind of method suffers from poor generalization to unseen data due to its strong dependence on training data, while acquiring high-quality rigging data is prohibitively expensive. To address this challenging problem, we propose G-Skin, a novel generative skinning framework designed for expressive and high-fidelity animation with 3D Gaussian representation. To overcome this 3D data scarcity, we introduce a skeleton-controllable image generation model leveraging 2D vision foundation models to distill powerful motion priors into pseudo-guidance. Guided by these priors, we formulate an optimization pipeline incorporating geometry-aware regularizations, which stabilizes the learning process and ensures smooth, structurally coherent skinning weights. G-Skin also generalizes flexibly to the augmented variants of 3D Gaussian representation designed to mitigate animation-induced rendering artifacts. Extensive experiments validate the effectiveness of our approach, demonstrating clear advantages over state-of-the-art methods. Project page: \url{https://yaoyx689.github.io/GSkin.html}.
\end{abstract}


\section{Introduction}

\begin{figure*}[ht] 
    \centering
    \includegraphics[width=1\textwidth]{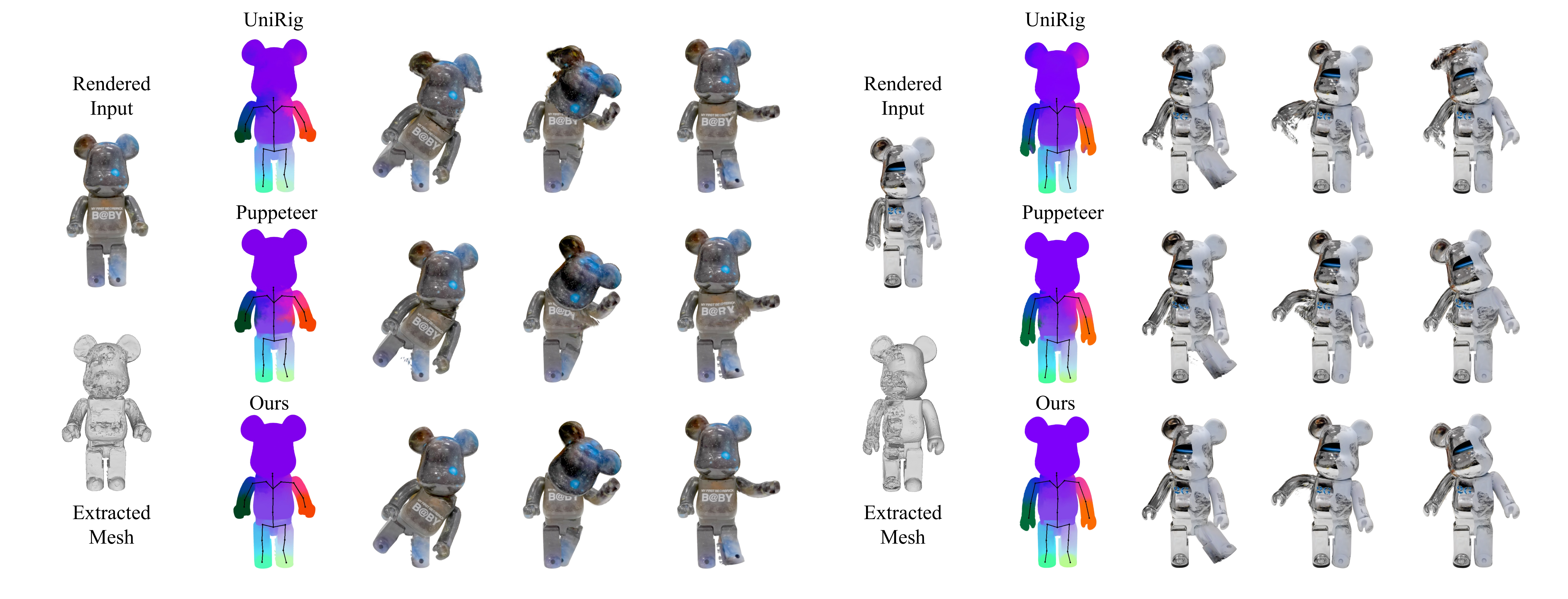}\vspace{-0.4cm} 
    \caption{\revise{Impact of extracted mesh quality on skinning. The left and right figures show two different examples featuring a similar object (a bear) with comparable motions, but distinct textures and materials. It can be clearly observed that the reconstructed mesh quality is significantly degraded in areas with specular reflections. This inconsistent mesh quality subsequently leads to divergent skinning weights, as seen in the left and right sides of the bear in the right figure, where mesh-based methods (i.e., UniRig~\cite{zhang2025unirig} and Puppeteer~\cite{song2025puppeteer}) yield visibly inconsistent skinning weights. In contrast, our method is much less affected. Here, the meshes used in UniRig and Puppeteer were extracted using GOF~\cite{Yu2024GOF}.}}
    \label{fig:bad_mesh} 
\end{figure*}

\IEEEPARstart{3}{D} Gaussian Splatting (3DGS) \cite{kerbl20233dgs} has revolutionized the field of novel view synthesis and 3D reconstruction. Driven by its explicit, flexible nature, 3DGS achieves unprecedented photorealistic rendering quality coupled with exceptional computational efficiency. Consequently, an increasing number of high-fidelity digital assets are now natively created and represented using 3D Gaussian primitives. Beyond static scenes, generating high-quality, animatable 3D content remains a cornerstone of modern computer graphics, with extensive applications spanning virtual reality, gaming, filmmaking, and telepresence. 
\revise{This motivates the development of effective skinning methods that enable skeletal deformation of static 3D Gaussian assets while preserving their visual quality.}

A straightforward solution to bind 3D Gaussian primitives with a given skeleton is training a feed-forward model directly with ground-truth data. However, this is infeasible due to the lack of Gaussian representation-based 3D assets paired with ground-truth rigs. An alternative strategy is to transfer 3D mesh-based methods to this domain, especially given the recent rapid advancements in mesh-based rigging for arbitrary skeleton topologies~\cite{xu2020rignet,song2025puppeteer,zhang2025unirig,liu2025riganything,zhang2026skintokens}. However, directly applying these techniques to 3D Gaussian skinning encounters the following two key limitations:
\begin{enumerate}
     \item 3D Gaussian primitives are not strictly confined to mesh surfaces: a large portion floats above or below mesh faces, or resides inside the object volume, introducing a distinct spatial discrepancy between Gaussian centers and the underlying mesh surface. \revise{Although existing mesh-based skinning methods can be adapted to 3D Gaussians through geometric approximations, the discrepancy between the two representations may compromise their skinning accuracy. Alternatively, extracting a mesh from 3D Gaussians introduces an additional dependency on mesh reconstruction quality, as illustrated in Fig.~\ref{fig:bad_mesh}}; 
   and 
    \item Existing mesh-based skinning methods~\cite{zhang2025unirig,song2025puppeteer} generalize poorly to unseen data, primarily due to their heavy reliance on training data. Currently, available training data predominantly originates from Objaverse/Objaverse-XL~\cite{deitke2023objaverse,deitke2023objaversexl}, which lacks both sufficient data quality and diversity in asset categories. Expanding such datasets with richer, high-quality data requires professional artists to dedicate substantial labor, making it extremely difficult to obtain at scale.
\end{enumerate}

To address this challenge, we propose \textit{G-Skin}, 
\revise{a framework that leverages generative priors to perform automatic skinning of 3D Gaussian assets given an aligned skeleton of arbitrary topology.}
By capitalizing on the rich motion priors embedded within vision foundation models, our framework requires only a small amount of data, thereby significantly reducing data dependency compared to existing methods~\cite{zhang2025unirig,song2025puppeteer} and enabling superior generalization performance.
Specifically, we first introduce a skeleton-controllable guidance generation model that employs the skeleton as control signals to synthesize guidance images of deformed shapes in skeleton-aligned poses.
Leveraging these generated images, we formulate an optimization objective to learn skinning weights. To account for potential inaccuracies in the generative guidance, we incorporate two geometric regularization terms that enhance training stability.  The synergy between powerful generative priors and our stable optimization strategy enables \textit{G-Skin} to achieve superior generalization across diverse and complex objects. 
Furthermore, our method can be flexibly applied to augmented variants of 3D Gaussian representation designed to minimize rendering artifacts during animation, such as a mesh-anchored Gaussian representation when the corresponding mesh is available. 
Extensive experiments validate the effectiveness of the proposed method.

In summary, our contributions are as follows.
\begin{itemize}
    \item  We introduce \textit{G-Skin}, a novel generative skinning framework for binding 3D Gaussian primitives to \revise{given} skeletons. Designed for high versatility, our framework can be flexibly applied to both standard discrete 3D Gaussian primitives and the artifact-minimizing variants.   
   \item We propose a skeleton-controllable generation model that distills motion priors from 2D foundation models to synthesize pose-aligned images, providing reliable guidance for automated skinning weight learning.
    \item We design a robust optimization strategy incorporating specialized regularization terms to mitigate inaccuracies in generative guidance, enabling stable skinning weight estimation and superior generalization across diverse, complex objects.
\end{itemize}

The rest of this paper is organized as follows. Sec.~\ref{sec:related} reviews the related work. Sec.~\ref{sec:proposed-method} details our proposed method. Sec.~\ref{sec:experiments} presents the experimental setup, evaluation, and analysis. Finally, Sec.~\ref{sec:conclusion} concludes the paper with a discussion on limitations and future work.

\section{Related Work}
\label{sec:related}

\subsection{3D Representation for Animated Objects}
Traditional pipelines typically rely on mesh-based representations, utilizing skeletons and skinning weights for deformation. While meshes excel in structural control, capturing photorealistic details of complex real-world objects often necessitates elaborate material modeling and high-resolution geometry, incurring substantial computational overhead. To achieve higher fidelity, some methods incorporate Neural Radiance Fields (NeRF)~\cite{mildenhall2020nerf} into animatable models~\cite{uzolas2023template,weng_humannerf_2022_cvpr}. However, NeRF-based techniques are often bottlenecked by high computational costs during both training and inference.

Recently, 3DGS~\cite{kerbl20233dgs} has emerged as a promising alternative, enabling real-time rendering with competitive visual quality. Nonetheless, the discrete nature of 3DGS makes it prone to artifacts during deformation. To enhance animation and editing stability, some methods~\cite{gao2024mesh,shao2024splattingavatar,qian2024gaussianavatars} anchor 3D Gaussian primitives to mesh facets to exploit geometric continuity. Similarly, VR-GS~\cite{jiang2024vr} embeds 3D Gaussian primitives into tetrahedra for piecewise linear deformation, while ARAP-GS~\cite{han2025arap} employs local rigidity constraints. For articulated objects, several works~\cite{wan2024template,moreau2024human} compute deformed Gaussian covariances via Linear Blend Skinning (LBS). Notably, some approaches~\cite{hu2024gaussianavatar,yao2025riggs,wang2026gaussianimate} adopt isotropic 3D Gaussian primitives to mitigate deformation-induced artifacts, though this often comes at the expense of rendering fine-grained details.

\subsection{Skeleton Generation}
Generating skeletons semantically for static objects is a challenging task. It requires accurate positioning of joint points and establishing reasonable topological relationships. Some methods utilize shape priors to manually design a general skeleton for a class of objects, such as SMPL~\cite{loper2015smpl} for human bodies and SMAL~\cite{zuffi2017smal} for quadrupeds. Since a wide range of shapes cannot be uniformly defined in advance, data-driven methods have been proposed to solve this problem. 
\cite{xu2019predicting} converts the input 3D shape into a voxel representation and predicts the skeleton by combining geometric shape features. RigNet~\cite{xu2020rignet} predicts the positions of joints based on a graph neural network and attention-driven clustering, and proposes BoneNet to predict the probability of links between joint points to obtain a complete skeleton. DRiVE~\cite{sun2025drive} proposes a 3D Gaussian-based diffusion network to accurately predict joint positions. Anymate~\cite{deng2025anymate} contributes a big rigging dataset including multi-category objects, and tests regression-based, diffusion-based, and volume-based architectures for joint prediction. Recently, with the popularity of the autoregressive model, some methods~\cite{liu2025riganything,zhang2025unirig,sun2025armo,song2025magicarti,song2025puppeteer} adopt it to predict skeletons. 

\begin{figure*}[h] 
    \centering
    \includegraphics[width=1\textwidth]{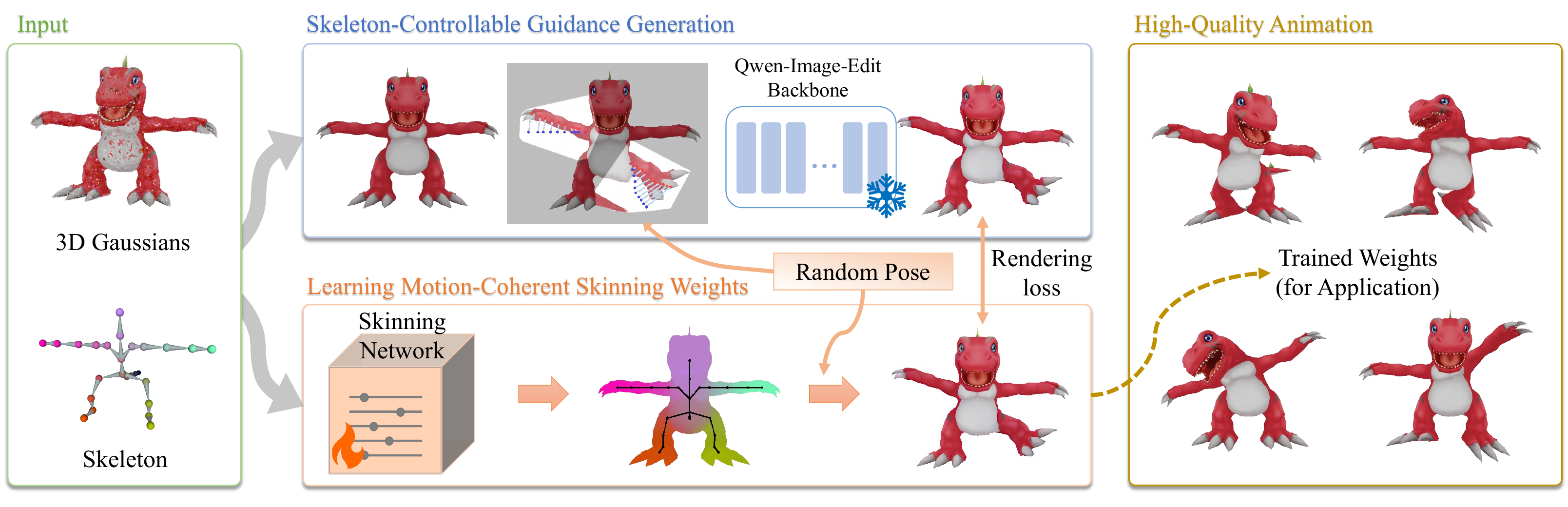}\vspace{-0.4cm}  
    \caption{\revise{Overview of our G-Skin. Given a 3D Gaussian representation and its corresponding skeleton, a skeleton-controllable guidance generation module is first introduced to leverage generative priors for producing guidance images. Subsequently, guided by these images, a robust learning framework incorporating two geometric regularization terms is designed to learn motion-coherent skinning weights. The optimized skinning weights are then used to drive high-quality 3DGS animations.}}
    \label{fig:pipeline} 
\end{figure*}

\subsection{Skinning Weight Learning}
Unlike sparse, intuitive, and potentially predefined skeletons, automated prediction of skinning weights would be more practical. 
Early works for automated skinning primarily focus on geometric heuristics, such as Laplace diffusion~\cite{Baran2007automatic} and volumetric geodesic distances~\cite{dionne2013geodesic}. With the advent of deep learning, several methods~\cite{liu2019neuro,mosella2022skinningnet} employ Graph Convolutional Networks (GCNs) to predict skinning weights by capturing mesh-skeleton topologies. More recently, attention mechanisms~\cite{chu2025humanrig,zhang2025unirig,song2025puppeteer} and diffusion frameworks~\cite{song2025magicarti} have been introduced to further improve accuracy.

However, these techniques are predominantly mesh-based and their performance is often bottlenecked by the availability of high-quality 3D training data. When applied to 3D Gaussian Splatting, existing solutions are sparse. For instance, MIA~\cite{guo2025make} treats Gaussian primitives merely as point clouds, failing to account for their anisotropic covariance during deformation. Furthermore, MIA is template-dependent, requiring a unified skeleton for training and inference, which limits its flexibility. Other video-based reconstruction methods~\cite{wan2024template,yao2025riggs,wang2026gaussianimate} enable rigging of category-agnostic objects, but their applicability is constrained by the motion range in the input videos. 

\subsection{Generative Priors for Animated Objects}
In contrast to the scarcity of dynamic 3D data, the abundance of 2D data has catalyzed the rapid evolution of vision foundation models. Recent research has begun incorporating these rich visual priors into the animation of 3D objects. For instance, \cite{li2025articulated} utilizes the CogVideoX-5B model~\cite{yang2024cogvideox} and Score Distillation Sampling (SDS) to generate text-driven 3D motions for Gaussian-based models. Puppeteer~\cite{song2025puppeteer} employs text-to-video models to provide motion guidance for animating rigged 3D characters. Similarly, AnimaX~\cite{huang2025animax} and Articulate3D~\cite{deb2025articulate3d} leverage multi-view rendered images and diffusion models to optimize 3D motion sequences or target poses through keypoint alignment and self-attention mechanisms.

While these methods successfully harness generative priors for pose estimation and motion synthesis, they primarily operate on pre-rigged models, leaving them dependent on pre-defined skinning weights and unable to address the rigging process itself. In contrast, we explore a novel direction of leveraging generative priors to guide the learning of skinning weights, effectively overcoming the limitations imposed by the scarcity of 3D supervision.

\section{Proposed Method}
\label{sec:proposed-method}

As illustrated in Fig.~\ref{fig:pipeline}, given a static object represented by 3D Gaussian primitives and the corresponding skeleton extracted through a typical off-the-shelf approach (e.g.,~\cite{guo2025make, song2025magicarti, song2025puppeteer, zhang2025unirig}) or manual construction, we propose a new pipeline to bind them, thereby facilitating realistic animation. 
Initially, we introduce the skeleton-related deformation representation (Sec.~\ref{sec:skeleton_rep}) and deformed 3D Gaussian representation (Sec.~\ref{sec:3dgs_rep}).
Following this, we optimize motion-coherent skinning weights by capitalizing on motion priors derived from a visual foundation model. To strictly maintain spatial consistency between the kinematic skeleton and the synthesized shape representation, we design a skeleton-controllable image generation scheme that leverages the 2D generative priors. 
By articulating the skeleton and automatically formulating control signals, we synthesize multi-view images with fine-grained spatial control, which subsequently act as tailored geometric guidance for the skinning weight learning (Sec.~\ref{sec:guidance_generation}). To stabilize the learning process against the inherent unreliability of generated images, we formulate an optimization objective that incorporates a local rigidity regularization term and a structural coupling term. This ensures the robust estimation of skinning weights, which in turn leads to motion-coherent and high-fidelity deformations of the 3D Gaussian primitives (Sec.~\ref{sec:learning-skinning}). 

\vspace{0.5em}
\mypara{Remark} Our method is designed \textit{primarily} for automated skinning of 3D Gaussian representation-based assets with arbitrary skeletons. By leveraging vision foundation models as priors, our approach reduces \revise{reliance on large-scale, high-quality 3D Gaussian rigging datasets} and significantly improves generalization to unseen shapes.

\subsection{Skeleton Representation and LBS-based Deformation}
\label{sec:skeleton_rep}
The skeleton is usually represented as a directed tree structure  $\mathcal{B}=\{\mathcal{J}, \mathcal{A}\}$, where $\mathcal{J}=\{\mathbf{J}_b\}$ denotes the set of joints, and $\mathcal{A} = \{A_b\}$ represents the directed connection relationships between joints.
Similar to~\cite{wu2023magicpony,uzolas2023template,yao2025riggs},
we define global translation ${\mathbf{t}}$ and rotational transformations $\{\mathbf{R}_{b}\}_{\mathbf{J}_b\in\mathcal{J}}$, representing the rotations relative to their parent joints. Here, the parent of the root joint $\mathbf{J}_{\text{root}}$ is itself.  
Without loss of generality, we define $\mathbf{J}_1$ as the root node.
Then a point $\mathbf{x}$ on the shape surface can be deformed via Linear Blend Skinning (LBS)~\cite{lewis2023pose}:
\begin{equation}
\label{eq:skeleton-deform}
\widehat{\mathbf{x}} = \sum_{b=1}^{|\mathcal{J}|} {\omega}_{\mathbf{x},b} \mathbf{P}_b \overline{\mathbf{x}}
\end{equation}
where
\begin{equation*}
\mathbf{P}_b=\mathbf{P}_{A_b} \hat{\mathbf{P}}_{b}, ~~\hat{\mathbf{P}}_b =
\begin{bmatrix}
\mathbf{R}_{b} & \mathbf{J}_{A_b} - \mathbf{R}_{b} \mathbf{J}_{A_b} \\
\mathbf{0} & 1
\end{bmatrix}
~\text{for } b \geq 2, 
\end{equation*}
\begin{equation}
\label{eq:skeleton_chain_deform}
\text{and}~~
\mathbf{P}_1=
\begin{bmatrix}
\mathbf{R}_1 & \mathbf{J}_1 - \mathbf{R}_1 \mathbf{J}_{1} + \mathbf{t} \\
\mathbf{0} & 1
\end{bmatrix}.
\end{equation}
Here $\mathbf{P}_b$ is defined recursively by its parent $\mathbf{P}_{A_b}$; $\overline{\mathbf{x}}$ is the homogeneous coordinate representation of $\mathbf{x}$;
${\omega}_{\mathbf{x},b}$ are the skinning weights. 
To simplify the representation, we denote the pose transformations of the skeleton as $\mathbf{P}=\{\mathbf{t},\{\mathbf{R}_b\}_{\mathbf{J}_b\in\mathcal{J}}\}$.

\subsection{Skeletal Deformed 3D Gaussian Representation}
\label{sec:3dgs_rep}
3DGS~\cite{kerbl20233dgs} represents 3D scenes or objects via a collection of anisotropic Gaussian primitives $\mathcal{G} = \{G_i: \boldsymbol{\mu}_i, \boldsymbol{\Sigma}_i, \sigma_i, sh_i\}$, with each primitive characterized by its mean position $\boldsymbol{\mu}_i$, opacity $\sigma_i$, and spherical harmonic coefficients $sh_i$. We denote the set of Gaussian centers by $\mathcal{C}=\{\boldsymbol{\mu}_i\}$.  
In contrast to existing category-agnostic mesh-based skinning approaches~\cite{zhang2025unirig,song2025puppeteer,xu2020rignet,liu2025riganything}, which are 
\revise{not natively designed for} discrete Gaussian primitives, our framework is primarily built for vanilla 3D Gaussian representation. Furthermore, it also generalizes flexibly to advanced variants developed to mitigate deformation artifacts during animation, such as widely adopted mesh-anchored 3D Gaussian representations~\cite{shao2024splattingavatar,qian2024gaussianavatars,gao2024mesh,waczynska2024games}. 

\vspace{0.5em}
\mypara{Vanilla 3D Gaussian Representation}
For the vanilla 3D Gaussian representation, we directly employ Eq.~\eqref{eq:skeleton-deform} to deform the Gaussian center $\boldsymbol{\mu}_i$ into $\hat{\boldsymbol{\mu}}_i$. The deformed covariance $\widehat{\boldsymbol{\Sigma}}_i$ is then estimated via:
\begin{equation}
\label{eq:3dgs_deform_cov}
\widehat{\boldsymbol{\Sigma}}_i = \widehat{\mathbf{T}}_i\boldsymbol{\Sigma}_i\widehat{\mathbf{T}}_i^T,
\end{equation}
where $\widehat{\mathbf{T}}_i$ is the rotation matrix converted from the blended quaternion $\widehat{\mathbf{q}}_i = \sum_{b}\omega_{\boldsymbol{\mu}_i,b}\mathbf{q}_b$, and $\mathbf{q}_b$ denotes the quaternion representation of the global rotation component of $\mathbf{P}_b$ in Eq.~\eqref{eq:skeleton_chain_deform}.

\vspace{0.5em}
\mypara{Mesh-anchored 3D Gaussian Representation}
For a proxy mesh $\mathcal{M} = \{\mathcal{V}, \mathcal{F}\}$ corresponding to the 3D Gaussians, 
each Gaussian $\boldsymbol{\mu}_i$ is parameterized relative to its nearest triangle $f\in\mathcal{F}$ using barycentric coordinates $(w_{f_0}^i, w_{f_1}^i, w_{f_2}^i)$. To account for Gaussians hovering near the surface, we formulate the center as:
\begin{equation}\label{eq:gaussian-pos}\boldsymbol{\mu}_i = w_{f_0}^i \mathbf{v}_{f_0} + w_{f_1}^i \mathbf{v}_{f_1} + w_{f_2}^i \mathbf{v}_{f_2} + \beta r \mathbf{n}_f,
\end{equation}
where $\mathbf{n}_f$ is the face normal, $r$ is the circumradius of the triangle, and $\beta \in [-0.5, 0.5]$ is a learnable offset. During deformation, the Gaussian’s covariance is transformed to maintain local geometric consistency. The deformed covariance $\widehat{\boldsymbol{\Sigma}}_i$ is computed as:
\begin{equation}
\label{eq:deform-covariance}
\begin{aligned}
\widehat{\boldsymbol{\Sigma}}_i = &\widetilde{\mathbf{T}}_i\boldsymbol{\Sigma}_i\widetilde{\mathbf{T}}_i^T, \quad \text{with}~~
\widetilde{\mathbf{T}}_i =  (\sum_{j=0}^{2}w_{f_j}^i \widehat{\mathbf{R}}_{f_j}) 
 (\sum_{j=0}^2w_{f_j}^i \widehat{\mathbf{S}}_{f_j}).
\end{aligned}
\end{equation}
where $\widehat{\mathbf{R}}_{f_j}$ and $\widehat{\mathbf{S}}_{f_j}$ denote the local rotation and shearing transformations of vertex $\mathbf{v}_{f_j}$, respectively, computed following the formulation in~\cite{gao2024mesh}.
This formulation leverages the inherent continuity of the underlying mesh to regularize covariance estimation, which significantly reduces rendering artifacts and enhances the visual coherence of the deformed Gaussians.

\subsection{Skeleton-Controllable Guidance Generation}
\label{sec:guidance_generation}
3D datasets with ground-truth skinning weights are \textit{scarce} and of \textit{low quality}. In contrast, 2D vision foundation models trained on large-scale data have demonstrated remarkable performance (e.g.,~\cite{klingai,jimengai2025}), especially in terms of motion rationality. We therefore aim to leverage their motion priors to facilitate the learning of skinning weights.

We propose a skeleton-controllable guidance generation scheme that ensures deformation consistency between the skeleton and the underlying shape. 
Specifically, we develop an automated control signal generation scheme that incorporates control points derived from the projected skeleton, along with a dynamic mask to distinguish dynamic regions from static ones. 
\revise{We integrate the encoded
control signals with input-image representations to guide the Qwen-Image-Edit diffusion transformer~\cite{qwen}. We build on the control architecture of TVEdit~\cite{xie2026TVEdit},
augment it with an additional mask encoder, and fine-tune the resulting control modules for our task while keeping the pretrained backbone frozen.  Appendix~\ref{app:guidance_model} provides further details.}

\vspace{0.5em}
\noindent\textbf{Remark.} 
A straightforward approach involves employing image-to-video generation models to synthesize motion sequences for static objects. However, these models typically produce single-view videos, and achieving view-consistent multi-view synthesis remains a formidable challenge. Furthermore, even with high-quality multi-view outputs, a precise alignment between skeletal poses and visual content is required before learning skinning weights; otherwise, geometric inconsistencies between the skeleton and the synthesized shape will occur. These hurdles significantly complicate the practical implementation of this pipeline.
Alternatively, one could utilize image-to-video models via Score Distillation Sampling (SDS). However, due to the stochastic nature of generative models, accurately optimizing skinning weights without fine-grained control signals remains extremely difficult.

\vspace{0.5em}
\noindent\textbf{Initial Skinning Weights.}
For each bone $b$ connecting joint $\mathbf{J}_b$ and its parent $\mathbf{J}_{A_b}$, we compute the Euclidean distance between mesh vertex $\mathbf{v}\in\mathcal{V}$ and its closest point $\mathbf{x}_{\mathbf{v}}^b$ on bone $b$. To address the limitations of Euclidean distance, specifically its inability to account for geodesic separation (e.g., adjacent but disconnected regions like two feet), we incorporate a visibility-aware correction mechanism. Specifically, we cast a ray between $\mathbf{v}$ and $\mathbf{x}_{\mathbf{v}}^b$ and project it onto multi-view planes. If the projected segment intersects regions outside the object mask, the vertex and bone are deemed mutually invisible ($\text{Vis}(\mathbf{v},b)=0$). In such cases, a distance penalty $\delta$ is applied:
\begin{equation}
D_{\mathbf{v}, b}=\left\{
\begin{aligned}
&\|\mathbf{v}-\mathbf{x}_{\mathbf{v}}^b\|_2+\delta, &\quad \text{Vis}(\mathbf{v},b)=0,\\
&\|\mathbf{v}-\mathbf{x}_{\mathbf{v}}^b\|_2, &\quad \text{Otherwise}.
\end{aligned}
\right.
\end{equation}
where $\delta$ is empirically set to $10\%$ of the object's bounding box diagonal. The initial skinning weights are then formulated following~\cite{dionne2013geodesic}:
\begin{equation}
\label{eq:init_skin}
\tilde{\omega}_{\mathbf{v},b}^{\text{init}} = \left(\frac{1}{(1-\gamma) D_{\mathbf{v},b} + \gamma D_{\mathbf{v},b}^2}\right)^2,
\end{equation}
where $\gamma \in [0,1]$ controls the binding smoothness (default $\gamma=0.5$).  For vanilla 3DGS, we directly replace $\mathbf{v}$ with $\boldsymbol{\mu}\in\mathcal{C}$ to calculate the initial skinning weights $\tilde{\omega}_{\boldsymbol{\mu}, b}^{\text{init}}$. These weights are further normalized across bones.

\vspace{0.5em}
\noindent\textbf{Construction of Control Signals.} 
Since the skeleton of the static object is already available, we can generate new poses by assigning updated pose parameters. The original and transformed positions of the skeletal joints are then used as handle points and target points, respectively, to guide the generation of posed images. 

Specifically, we first select a rendered image $I_l \in \mathcal{I}$ with camera parameters $C_l$. For a set of randomly selected joints $\mathcal{J}^S = \{\mathbf{J}_s\} \subset \mathcal{J} \setminus \{\mathbf{J}_{\text{root}}\}$, we construct rotation matrices $\{\mathbf{R}^s_{(l, \theta_s)}\}$ via axis-angle representation. To ensure that joint transformations are clearly observable from the projected perspective, we set the rotation axes perpendicular to the image plane and randomly sample the rotation angles $\{\theta_s\}$. By assigning these rotation matrices to the selected joints while maintaining identity matrices for the others, we construct a set of pose parameters $\mathbf{P}_S$. The resulting deformed joint positions are denoted as $\mathbf{J}^{\mathbf{P}_S}$. 
Utilizing the camera parameters $C_l$, we project the skeletal joints in both the rest pose $\mathbf{P}_{\text{rest}}$ and the new pose $\mathbf{P}_S$ onto the 2D plane, yielding projected point sets $\{\overline{\mathbf{J}}^{\mathbf{P}_{\text{rest}}}\}$ and $\{\overline{\mathbf{J}}^{\mathbf{P}_S}\}$. Joints exhibiting positional displacements between these two sets are identified and designated as handle and target points for pose $\mathbf{P}_S$. The collection $\mathcal{P} = \{\mathbf{P}_S\}$ thus constitutes the set of candidate skeleton poses used for guided image generation. As illustrated in Fig.~\ref{fig:pipeline} and Fig.~\ref{fig:drag_condition}, the blue and red dots represent handle points and target points, respectively.

\vspace{0.5em}
\noindent\textbf{Dynamic Mask Construction.}
The dynamic mask is essential for distinguishing regions requiring deformation from those that should remain stationary. To this end, we first estimate coarse skinning weights based on the spatial relationship between the mesh surface and the skeleton. 
By applying the initial skinning weights in Eq.~\eqref{eq:init_skin} via LBS in Eq.~\eqref{eq:skeleton-deform}, 
we obtain the deformed vertices $\widehat{\mathcal{V}}^{\text{init}}_{\mathbf{P}_S}$, which are subsequently rendered to produce the initial pose-transformed image $\hat{I}^{\text{init}}_{\mathbf{P}_{S}}$. The preliminary dynamic mask is derived from the pixel-wise difference between the reference image $I_{l}$ and the rendered image $\hat{I}^{\text{init}}_{\mathbf{P}_{S}}$. To ensure spatial coherence, we compute the convex hull for each connected component in the difference map to generate the final dynamic mask $M_{\mathbf{P}_S}$ (see Fig.~\ref{fig:pipeline} and Fig.~\ref{fig:drag_condition}). This mask, along with the designated handle and target points, is encoded into motion guidance signals to steer the image generation process.

\subsection{Learning Motion-Coherent Skinning Weights}
\label{sec:learning-skinning}
\mypara{Network Structure}
We adopt a CNN-based architecture~\cite{weng_humannerf_2022_cvpr} that leverages explicit volumetric features to model skinning weights.
Let $F_{\Phi}(\cdot)$ be the CNN parameterized by $\Phi$. 
Compared to directly optimizing point-wise skinning weights, this architecture exhibits superior stability owing to its local smoothness. Rather than directly regressing point-wise weights via an MLP, this approach inherently promotes stronger spatial locality.
We parameterize the canonical skinning field as a \revise{volumetric field} decoded from a latent embedding $\mathbf{z}$.
\revise{For efficiency, we use a query-pruned implementation of the dense CNN decoder: we retain a dense coarse grid of resolution $32^3$ and sparsely evaluate the finer decoder stages required for querying the canonical vertices.}
To query the skinning weights for \revise{a canonical vertex} $\mathbf{y} \in \mathbb{R}^3$ \revise{within the canonical bounding box}, we first transform its coordinates into a normalized coordinate system.
Given \revise{the axis-aligned bounding box of the canonical geometry}, defined by $[\mathbf{y}_{\min}, \mathbf{y}_{\max}]$, the normalized coordinate $\tilde{\mathbf{y}}$ is computed via an affine transformation:
\begin{equation*}
\tilde{\mathbf{y}} = (\mathbf{y} - \mathbf{y}_{\min}) \cdot \frac{2.0}{\max(\mathbf{y}_{\max} - \mathbf{y}_{\min})} - 1.0.
\end{equation*}
\revise{The maximum is taken over the three coordinate axes, giving a uniform scale factor that preserves the geometry's aspect ratio. This mapping aligns the minimum corner of the bounding box with $(-1,-1,-1)$ and maps the longest dimension to $[-1,1]$. The shorter dimensions are not additionally centered; the remaining space lies on the positive side of each corresponding axis. The geometry is therefore contained within the $[-1,1]^3$ sampling cube without additional bounding-box padding.}
We then retrieve the continuous weight vector $\mathbf{w}(\mathbf{y})$ through trilinear interpolation over the decoded volume\revise{, with the sampling-cube boundaries aligned with the centers of the boundary voxels}.
The voxel resolution is set to \revise{$256^3$, using the sparse evaluation described above,} in our experiments.

\vspace{0.5em}
\mypara{Initialization} During per-asset optimization, we first initialize the network with the geometry-based skinning weights in Eq.~\eqref{eq:init_skin}. That is to optimize the $\ell_2$-norm between the initial skinning weights and predicted skinning weights $\omega_{\mathbf{x},b}^{\text{learn}}$:
\begin{equation}
L_{\text{init}}=\sum_{\mathbf{x}}\sum_{b}\|\omega_{\mathbf{x},b}^{\text{learn}}-\tilde{\omega}_{\mathbf{x},b}^{\text{init}}\|^2.
\end{equation}
Here and in the following, $\mathbf{x}$ denotes $\mathbf{v}\in\mathcal{V}$ or $\boldsymbol{\mu}\in\mathcal{C}$. 

\vspace{0.5em}
\mypara{Learning with Motion Prior} We randomly sample a subset of poses from $\mathcal{P}$. 
For each skeleton pose $\mathbf{P}_{S}$, given its corresponding guidance image $\hat{I}_{\mathbf{P}_S}^{\text{gen}}$ and deformed mesh vertices $\widehat{\mathcal{V}}_{\mathbf{P}_S}$, we minimize the following loss function: 
\begin{equation}
\label{eq:motion-prior-loss}
L_{\mathbf{P}_S} = L_{\text{render}} 
 + w_{\text{a}}L_{\text{ARAP}} + w_{\text{c}}L_{\text{couple}},
\end{equation}
where \revise{$L_{\mathrm{render}}=0.8\ell_1+0.2(1-L_{\mathrm{SSIM}})$} comprises a weighted combination of $\ell_1$ and D-SSIM losses between the guidance image $\hat{I}^{\text{gen}}_{\mathbf{P}_S}$ and the rendered image $\hat{I}_{\mathbf{P}_S}$. $w_{\text{a}}$ and $w_{\text{c}}$ are weighting factors that balance the respective loss terms. \revise{We set \(w_a=10^3\) and \(w_c=0.1\) by default.}  

To ensure the stable optimization of skinning weights and mitigate the impact of minor artifacts in the guidance images, we introduce two regularization terms. First, the As-Rigid-As-Possible (ARAP) loss $L_{\text{ARAP}}$ preserves the local rigidity and smoothness of the deformed surface, which is defined as:
\begin{equation}
\label{eq:arap_loss}
L_{\text{ARAP}} = \frac{1}{2|\mathcal{E}|}\sum_{\mathbf{x}_j}\sum_{\mathbf{x}_k\in\mathcal{N}(\mathbf{x}_j)}\|(\widehat{\mathbf{x}}_k-\widehat{\mathbf{x}}_j) - \overline{\mathbf{R}}_j ({\mathbf{x}}_k-{\mathbf{x}}_j) \|^2,
\end{equation}
where $\overline{\mathbf{R}}_j\in SO(3)$ is an optimal local rotation matrix, and $|\mathcal{E}|$ denotes the number of edges (or neighbors). If a mesh is available, $\mathbf{x}_j$ denotes $\mathbf{v}_j\in\mathcal{V}$ and $\mathcal{N}(\cdot)$ is defined by the mesh connectivity. Otherwise, $\mathbf{x}_j$ denotes $\boldsymbol{\mu}_j\in\mathcal{C}$ and $\mathcal{N}(\cdot)$ represents the $K_1$-nearest neighbors, where we set $K_1=30$ by default.

Second, the coupling loss $L_{\text{couple}}$ is designed to maintain consistent relative distances between the skeleton and its proximal surface points during animation. Specifically, we uniformly sample points $\{\mathbf{u}\}$ on the skeletal bones and identify their $K_2$ nearest neighbors $\{\mathbf{x}_{\mathbf{u}}^{k}\}_{k=1}^{K_2}$ on the rest-pose surface. \revise{We set $K_2=6$ by default.} The coupling loss is then formulated as:
\begin{equation}
\begin{aligned}
L_{\text{couple}}
= \frac{1}{K_2|\{\mathbf{u}\}|}
\sum_{\mathbf{u}}\sum_{k=1}^{K_2}
\revise{
\left(
\frac{
\|\mathbf{u}-\mathbf{x}_{\mathbf{u}}^{k}\|
-
\|\hat{\mathbf{u}}-\hat{\mathbf{x}}_{\mathbf{u}}^{k}\|
}{
\max\!\left(
\|\mathbf{u}-\mathbf{x}_{\mathbf{u}}^{k}\|,\epsilon
\right)
}
\right)^2
}.
\end{aligned}
\end{equation}
Here $\hat{\mathbf{u}}$ and $\hat{\mathbf{x}}_{\mathbf{u}}^{k}$ denote the deformed positions of $\mathbf{u}$ and ${\mathbf{x}}_\mathbf{u}^{k}$ under pose $\mathbf{P}_S$, respectively, where $\mathbf{P}_S$ is omitted for notational brevity.  \revise{We normalize the distance change by the corresponding
rest-pose distance to penalize relative geometric distortion.
$\epsilon$ is a scale-dependent stabilizer set to half the median rest-pose edge length,
using mesh edges or 10-NN edges
between Gaussian centers for pure 3DGS.}

Finally, the learned skinning weights $\{\omega_{\mathbf{x},b}^{\text{learn}}\}$ are obtained, which are used to determine the skeleton-controllable deformations of 3DGS according to Eq.~\eqref{eq:3dgs_deform_cov} or Eqs.~\eqref{eq:gaussian-pos} and \eqref{eq:deform-covariance}. 

\section{Experiments}
\label{sec:experiments}

In this section, we first introduce the experimental setup, including our newly constructed 3D Gaussian representation-based assets for evaluation and the implementation details (Sec.~\ref{sec:exp_setup}). 
\revise{Next, we compared our method with existing mesh-based methods using mesh-anchored Gaussians on generated 3D assets, publicly available rigged datasets, and real-world data (Sec.~\ref{sec:compare_meshgs}). Furthermore, we evaluated our method on vanilla 3D Gaussian representation (Sec.~\ref{sec:compare_3dgs}).}
Finally, we present ablation studies and discussions (Sec.~\ref{sec:ablations}).

\subsection{Experiment Settings}
\label{sec:exp_setup}

\noindent\textbf{Generated 3D Assets.} 
Since \textbf{\textit{no existing}} datasets are specifically tailored for 3D Gaussian rigging, and most available 3D assets are either restricted to narrow categories (e.g., human bodies) or suffer from coarse geometric quality (e.g., Articulation XL 2.0~\cite{song2025magicarti}), we curated a new benchmark dataset for comprehensive evaluation.
We constructed a challenging evaluation set by generating novel objects using text-to-image models such as Flux~\cite{labs2025flux1kontextflowmatching} and Doubao AI~\cite{doubao2024}, and subsequently converting them into 3D assets using Trellis~\cite{xiang2025structured}. 
This process aligns closely with the practical needs of content creators. It includes 50 human-like and 35 non-humanoid assets, each comprising a complete 3D mesh and its corresponding 3D Gaussian representation. The latter serves as the primary input for our framework. 
To rigorously assess the robustness of our method, these assets were intentionally designed to be non-standard, encompassing cartoon figures, plush toys, and humans in loose or accessorized clothing, while deliberately avoiding simplified, tight-fitting models.

\vspace{0.5em}
\noindent\textbf{Implementation Details.} 
Owing to the lack of ground-truth rigging of 3D Gaussian primitives, we employed the mesh-based dataset to construct the training dataset. 
We selected 351 objects from the Articulation-XL 2.0 dataset~\cite{song2025magicarti} and enriched them by retrieving corresponding raw data from Objaverse-XL~\cite{deitke2023objaverse, deitke2023objaversexl} to extract their texture information.
We then synthesized approximately 100k image pairs by applying random deformations and rendering the objects from four distinct camera views. Instead of training from scratch, we performed fine-tuning to leverage and preserve the robust 2D generative priors of the foundation model, while ensuring that the synthesized outputs are structurally aligned with our skeletal guidance signals. Crucially, since the pre-trained 2D foundation model already possessed rich representations of object geometry and semantics, our fine-tuning process focused solely on aligning these shapes with the target skeletal poses. Therefore, the training data is sufficient to achieve high-quality results without causing prior degradation (or corrupting the pre-trained priors).

We selected four viewpoints (front, back, front-left, and front-right) to generate the guidance images. Specifically, \revise{25} images were synthesized per viewpoint, resulting in a total of \revise{100} guidance images for the skinning optimization phase. For each viewpoint, we synthesized random poses by applying rotation angles ranging from $0.1\pi$ to $0.33\pi$ to 10\% of randomly sampled joints, excluding the root joint. This configuration was employed for both training the image generation model and optimizing skinning weights. 
\revise{The Qwen-Image-Edit-based skeleton-controllable image generation model was trained on two Huawei Ascend 910B GPUs for 5 epochs with a batch size of 2 and a learning rate annealed from \(10^{-5}\) to \(10^{-6}\). We set the number of diffusion inference steps to 3.}
The optimization was conducted using the Adam optimizer~\cite{diederik2014adam} with a batch size of 20 for \revise{200} %
iterations. %
Unless otherwise specified, all experiments were conducted on a single GPU with 48 GB of VRAM. All images were rendered at a $512\times 512$ resolution.

\begin{figure*}[t] 
    \centering
    \includegraphics[width=1\textwidth]{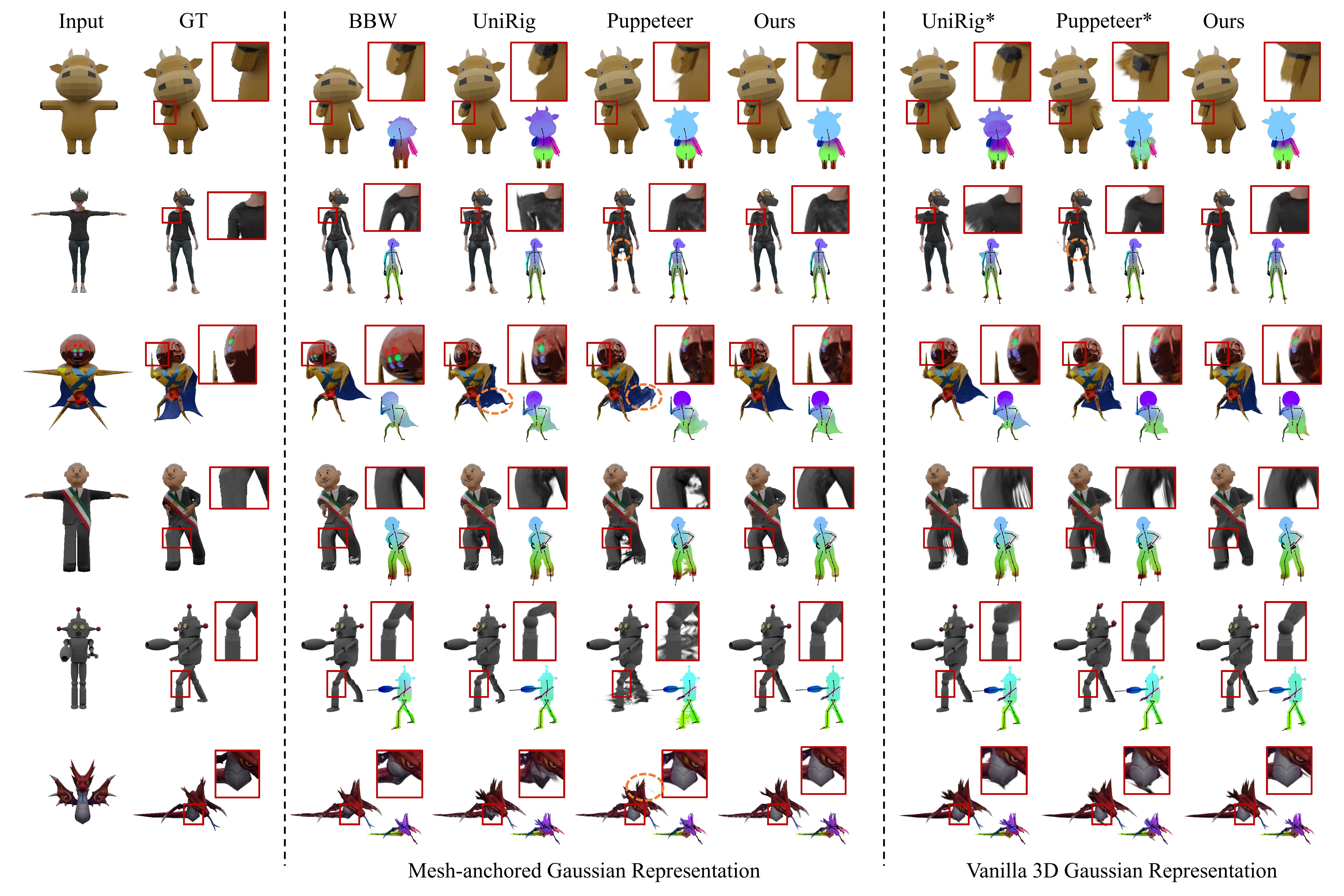} \vspace{-0.6cm} 
    \caption{
    \revise{Qualitative comparison on the Articulation-XL 2.0 dataset for both mesh-anchored and vanilla 3D Gaussian representations. It also displays both the deformed shape and the input rest-pose shape.}
    }
    \label{fig:compare_obj_rest} 
\end{figure*}

\begin{figure}[ht] 
    \centering
    \includegraphics[width=1\columnwidth]{figs/compare_non_rest.pdf} 
    \vspace{-0.9cm}
    \caption{\revise{Qualitative comparison of skinning results for assets bound in non-rest poses. We use the first frame of each animation sequence as the bind pose. BBW fails on two examples, which are marked as ``No result''.}
    \vspace{-0.3cm}
    }\label{fig:compare_obj_nonrest} 
\end{figure}

\begin{figure*}[ht] 
    \centering
    \includegraphics[width=0.9\textwidth]{figs/compare_gene_less.pdf}
    \vspace{-0.5cm}
    \caption{
    \revise{Visual comparisons with the state-of-the-art methods on the generative humanoid (left) and non-humanoid (right) 3D assets.} }
    \label{fig:compare_generate} 
\end{figure*}

\subsection{Comparisons on Mesh-anchored Gaussian Representation}
\label{sec:compare_meshgs}

\mypara{\revise{Publicly Available Synthetic Data}}
To evaluate our method quantitatively against ground-truth skinning, we collected 40 diverse objects with associated motion sequences and textures from the test set of Articulation-XL 2.0~\cite{song2025magicarti}, and compared our approach with mesh-based baselines, including learning-based methods (UniRig~\cite{zhang2025unirig} and Puppeteer~\cite{song2025puppeteer}) as well as an optimization-based method (BBW~\cite{jacobson2011bounded}).
We converted all target objects into mesh-anchored Gaussian representations. For a fair comparison, all methods used the same ground-truth skeletons for skinning and were rendered using the corresponding mesh-anchored 3D Gaussians.
For each object, we uniformly sampled 24 frames from the corresponding motion sequences, rendered each frame from 6 diverse viewpoints, and measured the rendering quality against ground-truth images using PSNR, SSIM, and LPIPS.
\revise{In addition to rendering quality, we further assess the geometric quality of skinning. We report the mean squared error (MSE) and maximum Euclidean distance ($D_{\text{Max}}$) between corresponding vertices of the meshes deformed using the predicted and ground-truth (GT) skinning weights, as well as the Chamfer distance (CD) between the resulting surfaces. %
Additionally, we report the L1 error between the predicted and GT skinning weights. } %

\revise{We evaluate skinning quality under two settings: binding the asset in its rest pose and binding it in a non-rest pose taken from the first frame of the animation sequence. As shown in Table~\ref{Tab:compare_obj} and Figs.~\ref{fig:compare_obj_rest} and~\ref{fig:compare_obj_nonrest}, our method achieves high rendering accuracy in both settings and yields the best results on nearly all geometric metrics.}
\revise{In Fig.~\ref{fig:compare_obj_nonrest}, our method outperforms existing approaches on assets in non-rest poses, even when the arms are in close contact with the torso. }
\revise{Our L1 error is higher than that of Puppeteer because our method relies on generative priors and does not explicitly enforce fine-grained agreement with the ground-truth skinning weights. Furthermore, 
objects in the \textbf{Articulation-XL 2.0 test set exhibit some similarities to those in the training sets of UniRig and Puppeteer}.} In some cases, identical assets (e.g., the Spiderman model) appear in both sets, which artificially boosts the performance of UniRig and Puppeteer and narrows the numerical gap with our method. Consequently, this data overlap prevents a clear evaluation of the true generalization capabilities of existing approaches.

\begin{table}[t]
	\caption{\revise{Comparison of the average PSNR, SSIM, LPIPS, MSE($\times 10^{-4}$), $D_{\text{Max}}$, CD($\times 10^{-3}$), L1 on the Articulation XL 2.0 dataset. ``$\uparrow$" (resp. $\downarrow$) indicates the larger (resp. smaller), the better. The best results are highlighted in bold.} 
    }\vspace{-0.25cm}
	\label{Tab:compare_obj}
    
		\setlength{\tabcolsep}{3.1pt}
	\centering
	\begin{small}
    \resizebox{\linewidth}{!}{
		\begin{tabular}{ l  c  c  c  c c c  c }
            \toprule 
             Method  & PSNR$^{\uparrow}$ & SSIM$^{\uparrow}$ & LPIPS$^{\downarrow}$ & MSE$^{\downarrow}$& $D_{\text{Max}}$$^{\downarrow}$& CD$^{\downarrow}$ & L1$^{\downarrow}$ \\
            \midrule
            \textit{i) rest} \\
            BBW~\cite{jacobson2011bounded}$^\dagger$ & 22.31 & 0.936 & 0.0571 & 5.42 & 0.19 & 9.20 & 1.01 \\
            UniRig~\cite{zhang2025unirig} & 25.61 & 0.959 & 0.0478 & 2.69 & 0.17 & 4.88 & 0.64 \\ 
            Puppeteer~\cite{song2025puppeteer} & 26.19 & 0.963 & 0.0475 & 2.74 & 0.21 & 4.28 & \textbf{0.38} \\
            Ours & \textbf{26.58} & \textbf{0.966} & \textbf{0.0426} & \textbf{1.93} & \textbf{0.14} & \textbf{4.02} & 0.51 \\
            \midrule 
            \textit{ii) non-rest} \\
            BBW~\cite{jacobson2011bounded}$^\dagger$ & 24.21 &	0.943 &	0.0575 & 4.66 &	0.22 &	6.82 &	0.99 \\ 
            UniRig~\cite{zhang2025unirig} & 25.18 &	0.953 &	0.0545 &	5.57&	0.28&	4.55&	0.76 \\%
            Puppeteer~\cite{song2025puppeteer} & 26.33 &	0.956 &	0.0564 &	{4.11}&	0.27&	\textbf{3.04} & \textbf{0.46} \\
            Ours & \textbf{26.65} &	\textbf{0.963} &	\textbf{0.0456} &	\textbf{3.91} &	\textbf{0.18} &	{3.16} & {0.63} \\ 
            \bottomrule
		\end{tabular}}
        \vspace{3pt}
    \begin{minipage}{\columnwidth}
        \footnotesize
        $^\dagger$For BBW, we use the implementation in libigl~\cite{libigl}. Due to its sensitivity to mesh quality, valid results could not be obtained for all cases despite remeshing. We report averages over the 29 successful cases in the rest setting and the 25 successful cases in the non-rest setting, excluding 11 and 15 cases, respectively.
    \end{minipage}
	\end{small}
    \vspace{-0.5cm}
\end{table}

\vspace{0.5em}
\mypara{\revise{Image-to-3D Generated Assets}}
To validate the generalization capability of our method, we conducted additional comparative evaluations on our newly generated 3D assets. 
This setup ensures that none of the baseline methods encountered these objects during training, thereby guaranteeing a fair, "unseen" benchmark that highlights the practical utility of our approach.
Specifically, we compared the proposed method against UniRig~\cite{zhang2025unirig} and Puppeteer~\cite{song2025puppeteer}, and a skeleton-fixed method, MIA~\cite{guo2025make} (restricted to humanoids). Due to its sensitivity to mesh quality and limited overall performance (as observed in Table~\ref{Tab:compare_obj} and Fig.~\ref{fig:compare_obj_rest}), BBW~\cite{jacobson2011bounded} was excluded from this set of experiments. \revise{For humanoid assets, we applied 20 Mixamo~\cite{mixamo} sequences to each asset, yielding 1,000 dynamic sequences. }
For non-humanoid assets lacking off-the-shelf driving sequences, we synthesized random poses for each asset and interpolated between them to create two continuous animation sequences for evaluation. \revise{These sequences were rendered from the front view into videos and evaluated using VBench~\cite{huang2023vbench}, a widely recognized video evaluation suite, across metrics including \revise{ subject consistency (Consis.) and aesthetic quality (Aes.).}
Quantitative results are presented in Table~\ref{Tab:generative_real_data}, and qualitative comparisons are illustrated in 
\revise{Fig.~\ref{fig:compare_generate}}. These evaluations demonstrate that our method achieves superior generalization and produces significantly fewer artifacts during animation, highlighting its robustness and practicality across diverse data sources. See Appendix~\ref{app:visualization} and the \textbf{project page} for more qualitative results.}

\vspace{0.5em}
\mypara{\revise{Real-World 3D Assets}}
\revise{We further evaluate our method on real-world captured data. We collected 24 objects suitable for animation from the OmniObject3D dataset~\cite{wu2023omniobject3d}, including teddy bears, toy animals, and dolls. We converted their reconstructed objects into mesh-anchored 3D Gaussian representations for our experiments. To facilitate skeletal posing for skinning evaluation, we divide these objects into humanoid (12 objects) and non-humanoid (12 objects) categories. For humanoid objects, we use Make-It-Animatable~\cite{guo2025make} to extract skeletons and animate them using motion sequences from Mixamo~\cite{mixamo}. For non-humanoid objects, we use Puppeteer~\cite{song2025puppeteer} to extract skeletons and generate random motion sequences. We construct two motion sequences for each object and evaluate the resulting animations using VBench~\cite{huang2023vbench}. The quantitative and qualitative results are presented in Table~\ref{Tab:generative_real_data} and Fig.~\ref{fig:real_data}, respectively. Our method also performs well, scoring slightly below MIA only on the Consis. metric.}

\begin{table}[t]
\centering
\caption{\revise{Comparison of animation video quality on generated and real-world assets.}}
\vspace{-0.25cm}
\setlength{\tabcolsep}{3.1pt}
\resizebox{\linewidth}{!}{%
\begin{tabular}{l | c c | c c}
    \toprule
    \multirow{2}{*}{Method}
    & \multicolumn{2}{c|}{Generative data}
    & \multicolumn{2}{c}{Real data} \\
    \cmidrule(lr){2-3}\cmidrule(lr){4-5}
    & Humanoid & Others & Humanoid & Others \\
    \midrule
    UniRig~\cite{zhang2025unirig}
    & 0.914 / 0.587
    & 0.951 / 0.612
    & 0.919 / 0.529
    & 0.941 / 0.483 \\
    Puppeteer~\cite{song2025puppeteer}
    & 0.917 / 0.599
    & 0.952 / 0.613
    & 0.948 / 0.563
    & 0.941 / 0.485 \\
    MIA~\cite{guo2025make}
    & 0.916 / 0.600
    & --
    & \textbf{0.949} / 0.570
    & -- \\
    Ours
    & \textbf{0.923} / \textbf{0.621}
    & \textbf{0.954} / \textbf{0.616}
    & 0.948 / \textbf{0.579}
    & \textbf{0.944} / \textbf{0.487} \\
    \bottomrule
\end{tabular}%
}
\begin{minipage}{\linewidth}
\vspace{2pt}
\footnotesize
Each entry reports VBench scores (Consis. / Aes.).
Higher values are better. The best results are highlighted in bold.
\end{minipage}
\label{Tab:generative_real_data}
\end{table}

\begin{figure*}[ht] 
    \centering
    \includegraphics[width=0.95\textwidth]{figs/compare_real.pdf}\vspace{-0.2cm}  
    \caption{\revise{Qualitative comparisons on OmniObject3D. Humanoid characters are shown on the left, and others are shown on the right.}
    }\label{fig:real_data} 
\end{figure*}

\begin{table}[t]
\centering
\caption{\revise{User-study results.}}
\label{Tab:user_study_pairwise}
\vspace{-0.25cm}
\setlength{\tabcolsep}{1.8pt}
\resizebox{\linewidth}{!}{
\begin{tabular}{l | cc | cc}
    \toprule
    \multirow{2}{*}{Ours vs.}
    & \multicolumn{2}{c|}{Generative data}
    & \multicolumn{2}{c}{Real data} \\
    \cmidrule(lr){2-3}\cmidrule(lr){4-5}
    & Humanoid & Others & Humanoid & Others \\
    \midrule
    UniRig~\cite{zhang2025unirig}
    & 97.3 / 1.9 / 0.8
    & 90.4 / 7.7 / 1.9
    & 98.8 / 0.8 / 0.4
    & 63.8 / 30.4 / 5.8 \\
    Puppeteer~\cite{song2025puppeteer}
    & 96.5 / 2.7 / 0.8
    & 79.6 / 15.4 / 5.0
    & 88.1 / 7.7 / 4.2
    & 68.5 / 28.5 / 3.1 \\
    MIA~\cite{guo2025make}
    & 99.2 / 0.8 / 0.0
    & --
    & 79.6 / 18.1 / 2.3
    & -- \\
    \bottomrule
\end{tabular}
}

\begin{minipage}{\linewidth}
\vspace{2pt}
\footnotesize
Each entry reports the percentages of judgments favoring our method,
indicating a tie, and favoring the competing method, respectively
(Ours / Tie / Competitor).
\end{minipage}
\end{table}

\vspace{0.5em}
\mypara{\revise{User Study}}
In addition to the objective metrics evaluated via VBench, we further conducted a user study to assess subjective visual performance. \revise{We selected 40 representative animation sequences: 20 from generated 3D assets (10 humanoid and 10 non-humanoid) and 20 from real-world 3D assets (10 humanoid and 10 non-humanoid).} 
\revise{We conducted a pairwise user study where each comparison presented Ours alongside one competing method, showing both rendered animations and skinning visualizations. The question order and left–right placement of the methods were randomized. Participants selected the result with better shape–skeleton compatibility and more plausible deformation. We collected 26 completed questionnaires and report the percentages favoring Ours, indicating a tie, and favoring the competing method. Table~\ref{Tab:user_study_pairwise} shows that our method achieves substantially higher win rates than the competing methods.}
\revise{Further analyses,} selected representative sequences, and additional visual comparisons are provided in Appendices~\ref{app:analysis} and~\ref{app:visualization} and on our \textbf{project page}.

\subsection{Comparisons on Vanilla 3D Gaussian Representation}
\label{sec:compare_3dgs}

We evaluated our method using the vanilla 3D Gaussian representation. 
\revise{Since no skinning model is directly available for 3D Gaussians with arbitrary skeletal topologies, we use the Gaussian centers in place of mesh vertices and estimate the normal at each center by applying PCA to its neighboring Gaussian centers. We then feed these centers and their estimated normals into the pretrained mesh-based models UniRig~\cite{zhang2025unirig} and Puppeteer~\cite{song2025puppeteer} to predict skinning weights. We denote these configurations by UniRig$^{*}$ and Puppeteer$^{*}$, respectively, to indicate that the models are applied to Gaussian data without retraining. We convert the Articulation-XL 2.0 test assets described in Sec.~\ref{sec:compare_meshgs} into vanilla 3D Gaussian representations and evaluate all methods on them. Quantitative and qualitative results are presented in Table~\ref{Tab:compare_obj_vanilla3dgs} and Fig.~\ref{fig:compare_obj_rest}, respectively.}

\begin{table}[t]
	\caption{\revise{Comparison on
Articulation-XL 2.0 using vanilla 3DGS.} 
    }\vspace{-0.25cm}
	\label{Tab:compare_obj_vanilla3dgs}
		\setlength{\tabcolsep}{3.1pt}
	\centering
	\begin{small}
		\begin{tabular}{ l  c  c  c }
            \toprule 
             Method  & PSNR$^{\uparrow}$ & SSIM$^{\uparrow}$ & LPIPS$^{\downarrow}$ \\
            \midrule
            UniRig$^*$~\cite{zhang2025unirig} & 25.32 & 0.959 & 0.0513  \\ 
            Puppeteer$^*$~\cite{song2025puppeteer} & 25.62 & 0.961 & 0.0509 \\
            Ours & \textbf{26.48} & \textbf{0.967} & \textbf{0.0462} \\
            \bottomrule
		\end{tabular}

\begin{minipage}{\linewidth}
\footnotesize
\vspace{2pt}
We report average PSNR, SSIM, and LPIPS.
$\uparrow$ and $\downarrow$ indicate that higher and lower values
are better, respectively. The best results are highlighted in bold.
\end{minipage}
	\end{small}
\end{table}

\revise{Since MIA~\cite{guo2025make} is designed to directly process discrete 3D primitives, we also include it in our comparisons.}
The pre-trained MIA model is restricted to human-like datasets, so we evaluated and compared both methods on our generated human-like 3D assets. Additionally, we evaluated our approach on isotropic variants of 3D Gaussian Splatting, following recent prior works~\cite{yao2025riggs,wang2026gaussianimate}.
\revise{Both methods used the same skeletons extracted by MIA. We applied Mixamo~\cite{mixamo} motion sequences to produce 1,000 animation sequences for evaluation.}
These sequences were rendered from the front view into videos and evaluated using VBench~\cite{huang2023vbench}, a widely recognized video evaluation suite, across metrics including subject consistency (Consis.) \revise{and} aesthetic quality (Aes.). %

\begin{figure}[ht] 
    \centering
    \includegraphics[width=1\columnwidth]{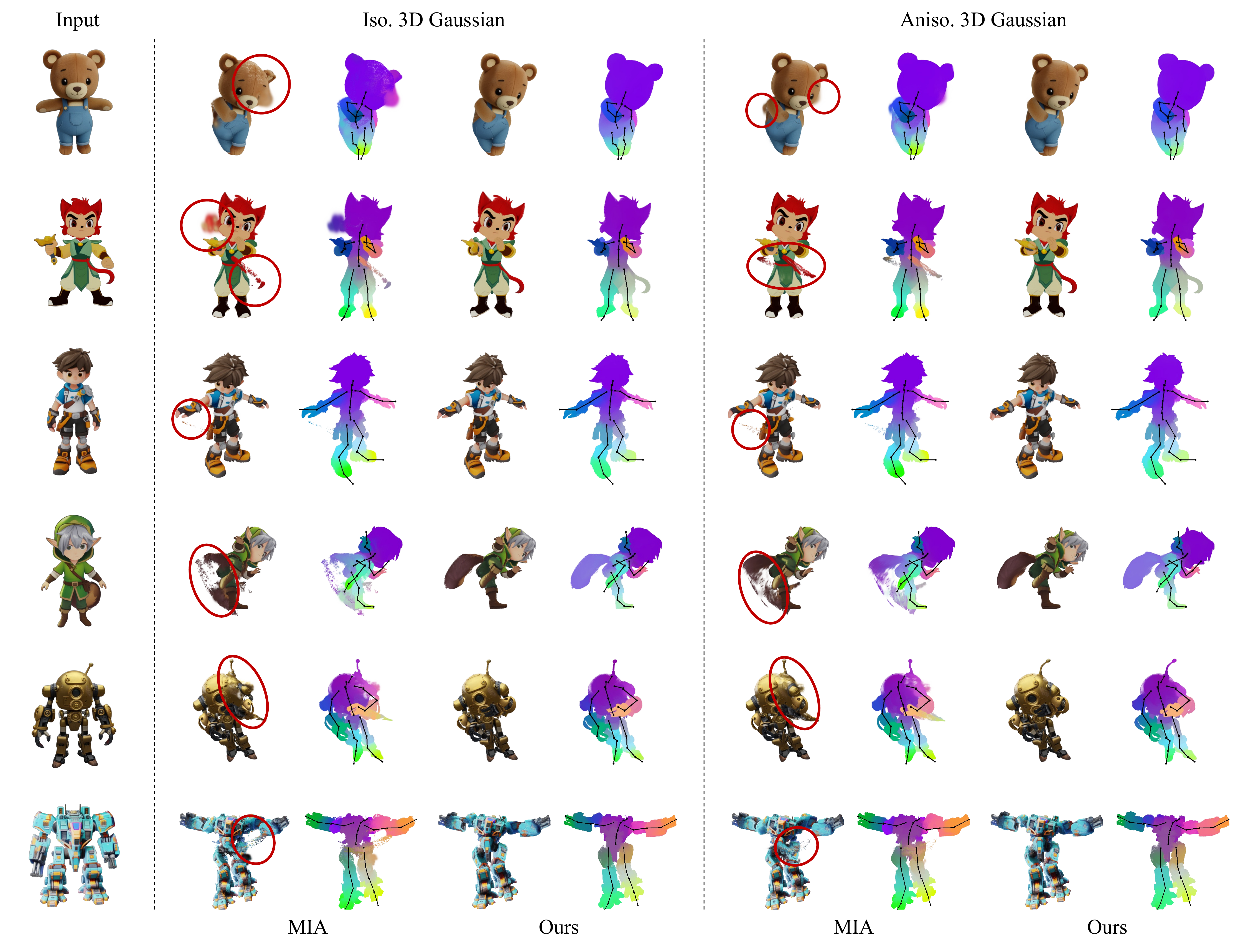} 
    \vspace{-0.5cm} 
    \caption{
    \revise{Visual comparison with MIA~\cite{guo2025make} on the generative 3D assets with isotropic/anisotropic 3D Gaussian representation.}}
    \label{fig:compare_3dgs} 
\end{figure}

The quantitative and qualitative comparisons illustrated in Fig.~\ref{fig:compare_3dgs} and Table~\ref{Tab:generative_data_3dgs} show that our method outperformed MIA. It is clearly demonstrated that our skinning field seamlessly generalizes to discrete Gaussian primitives, yielding highly compelling animation performance in both cases. 
More video results are available in the video demo on the \textbf{project page}.

\begin{table}[htb]
\centering
\caption{\revise{Comparison with MIA on generated assets using
isotropic and anisotropic vanilla 3DGS.}}
\vspace{-0.25cm}
\small
\setlength{\tabcolsep}{3pt}
\begin{tabular}{ l  c  c }
    \toprule
    Method & Iso. 3DGS & Aniso. 3DGS \\
    \midrule
    MIA~\cite{guo2025make}
    & 0.909 / 0.603 & 0.912 / 0.603 \\
    Ours
    & \textbf{0.924} / \textbf{0.622}
    & \textbf{0.924} / \textbf{0.618} \\
    \bottomrule
\end{tabular}

\begin{minipage}{\linewidth}
\vspace{2pt}
\footnotesize
Each entry reports VBench scores (Consis.~$\uparrow$ /
Aes.~$\uparrow$). ``$\uparrow$" (resp. $\downarrow$) indicates the larger (resp. smaller), the better. The best results are highlighted in bold.
\end{minipage}
\label{Tab:generative_data_3dgs}
\end{table}

\begin{table}[h]
\centering
\caption{\revise{Comparison of skinning performance across
different variants.}}\vspace{-0.25cm} 
\setlength{\tabcolsep}{3.1pt}
\resizebox{\linewidth}{!}{
\begin{small}
\begin{tabular}{ c  c  c  c c c c c}
            \toprule 
            Variant & PSNR$^{\uparrow}$ & SSIM$^{\uparrow}$ & LPIPS$^{\downarrow}$ & MSE$^{\downarrow}$ & $D_{\text{Max}}$$^{\downarrow}$ & CD$^{\downarrow}$ & L1$^{\downarrow}$ \\
            \midrule
            Init. skin. &  23.83 & 0.945 & 0.0559 & 3.97 & 0.20 & 6.83 & 0.78  \\
             w/o $L_{\text{ARAP}}$ & 26.16 & 0.962 & 0.0477 & 2.57 & 0.17  & 4.31 & 0.59 \\
             w/o $L_{\text{couple}}$ &  26.30 & 0.965 & 0.0433 & 2.07 & \textbf{0.14} & 4.20 & 0.57 \\
             Ours & \textbf{26.58} & \textbf{0.966} & \textbf{0.0426} & \textbf{1.93} & \textbf{0.14} & \textbf{4.02} & \textbf{0.51} \\
            \bottomrule
\end{tabular}
\end{small}}

\begin{minipage}{\linewidth}
\vspace{2pt}
\footnotesize
We report average PSNR, SSIM, LPIPS, MSE ($\times 10^{-4}$),
$D_{\text{Max}}$, CD ($\times 10^{-3}$), and L1 error.
$\uparrow$ and $\downarrow$ indicate that higher and lower values
are better, respectively. The best results are highlighted in bold.
\end{minipage}

\label{tab:ablas_optskin}
\end{table}

\subsection{Ablation Studies and Analysis} 
\label{sec:ablations}
\revise{In this subsection, we conducted ablation studies on individual components of our method and analyzed its sensitivity to skeleton quality and runtime. Further details and analyses are provided in Appendix~\ref{app:analysis}.}

\begin{figure*}[ht] 
    \centering
    \includegraphics[width=0.9\textwidth]{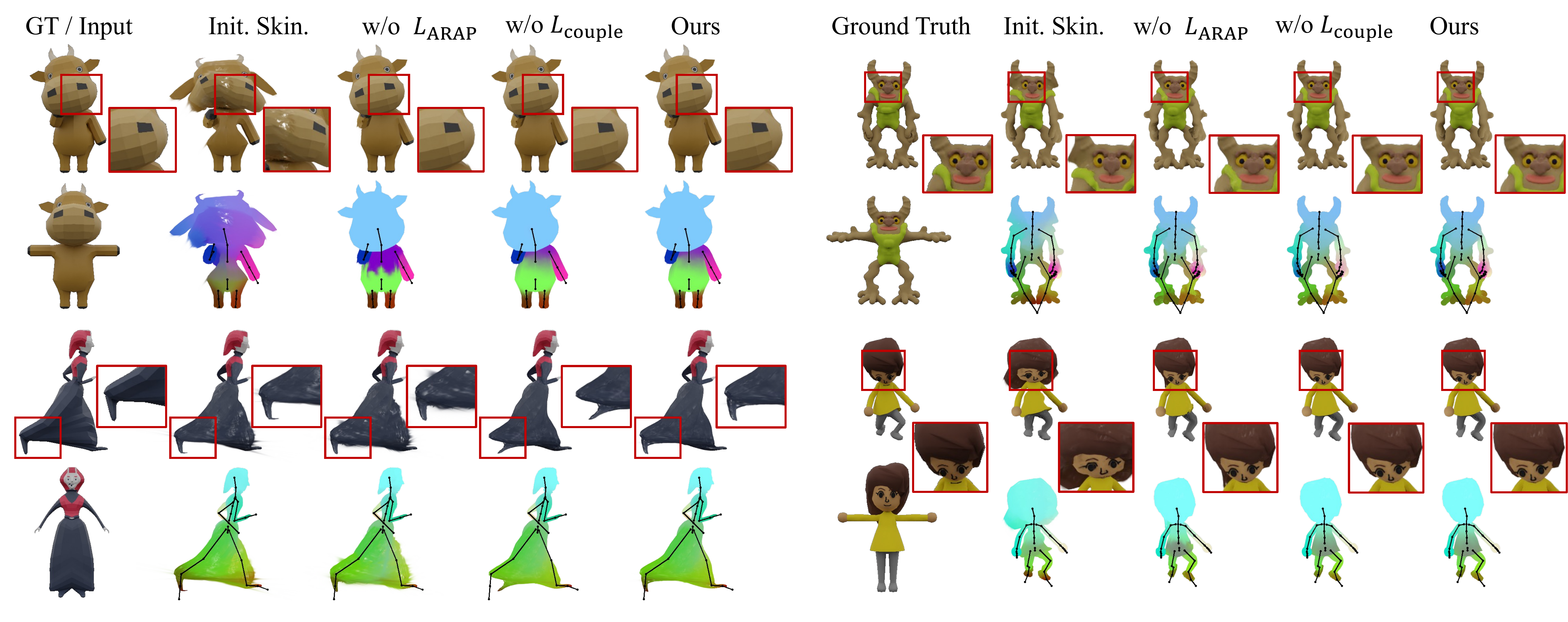} \vspace{-0.4cm} 
    \caption{
    \revise{Visual comparison of skinning optimization variants.}}
    \label{fig:ablas_loss} 
\end{figure*}

\vspace{0.5em}
\noindent\textbf{Optimization of Skinning Weights.}
To verify the effectiveness of each component of our method, we first evaluated the performance using only the initial skinning weights (Eq.~(\ref{eq:init_skin})) without further optimization (\textit{Init. skin.}). As shown in Table~\ref{tab:ablas_optskin}, this baseline leads to a notable degradation in accuracy, highlighting the necessity of our generative prior-guided optimization stage.  Furthermore, we conducted ablation studies to assess the impact of different loss terms during the motion-prior-guided skinning learning process. Specifically, we individually ablated $L_{\text{ARAP}}$ and $L_{\text{couple}}$ and show the quantitative and qualitative results in Table~\ref{tab:ablas_optskin} and Fig.~\ref{fig:ablas_loss}. They demonstrate that
$L_{\text{ARAP}}$ facilitates the local smoothness of the learned weights, and $L_{\text{couple}}$ effectively strengthens the geometric coupling between the skeleton and the object surface.

Furthermore, \revise{we also investigated the effect of the number of reference viewpoints on skinning weight learning by comparing configurations with 1 view (front), 2 views (front and back), and 6 views (front, back, front-left, front-right, left, and right), alongside our default 4-view configuration.
We also evaluate the effect of the number of guidance images per viewpoint by comparing settings with 13, 25, and 50 images per view while keeping the viewpoints fixed. Table~\ref{Tab:ablas_guidance} shows the numerical results.
Considering that generating each image takes 2.5 seconds, our default configuration of four views with 25 images per view provides a balance between optimization efficiency and animation quality.}

\begin{table}[h]
\centering
\caption{\revise{Effect of the number of guidance images and
viewpoints on skinning performance.}} \vspace{-0.25cm}
\setlength{\tabcolsep}{3.1pt}
\resizebox{\linewidth}{!}{
\begin{small}
\begin{tabular}{ c  c  c  c c c c c c}
            \toprule 
            Variant & PSNR$^{\uparrow}$ & SSIM$^{\uparrow}$ & LPIPS$^{\downarrow}$ & MSE$^{\downarrow}$ & $D_{\text{Max}}$$^{\downarrow}$ & CD$^{\downarrow}$ & L1$^{\downarrow}$ & \#P/J\\
            \midrule
            $25 \times 1$
            & 26.01 & 0.962 & 0.0451
            & 2.07 & 0.15 & 4.48 & 0.59 & 2.75 \\

            $25 \times 2$
            & 26.20 & 0.964 & 0.0439
            & 2.70 & 0.15 & 4.57 & 0.55 & 5.49 \\

            $13 \times 4$
            & 26.52 & 0.966 & 0.0428
            & 1.94 & 0.14 & 4.04 & 0.52 & 5.71 \\

            $25 \times 4$
            & 26.58 & 0.966 & 0.0426
            & 1.93 & {0.14} & 4.02 & 0.51 & 10.98 \\
             
            $50 \times 4$
            & {26.68}
            & {0.967}
            & {0.0423}
            & {1.92}
            & 0.14
            & {3.95}
            & {0.49}
            & 21.97 \\

            $25 \times 6$
            & {26.64}
            & {0.966}
            & {0.0425}
            & {1.93}
            & 0.14
            & {4.00}
            & {0.52}
            & 16.48 \\
            \bottomrule
\end{tabular}
\end{small}}

\begin{minipage}{\linewidth}
\vspace{2pt}
\footnotesize
$N \times V$ denotes $N$ guidance images per view across $V$ viewpoints.
\#P/J denotes the average number of guidance samples directly rotating
each non-root joint. MSE and CD are reported in units of $10^{-4}$
and $10^{-3}$, respectively. $\uparrow$ and $\downarrow$ indicate
that higher and lower values are better, respectively.
\end{minipage}
\label{Tab:ablas_guidance}
\end{table}

\begin{table}
\centering
\caption{\revise{Comparison of the average quality of generated images with different variants.}}\vspace{-0.25cm} 
\begin{small}
\begin{tabular}{ c  c  c  c}
            \toprule 
            Variant & PSNR$^{\uparrow}$ & SSIM$^{\uparrow}$ & LPIPS$^{\downarrow}$ \\ \midrule 
             w/o fine-tuning & 18.88 & 0.888 & 0.0890 \\
             25\% & 28.95 & 0.978 & 0.0212 \\
             50\% & 29.36 & 0.979 & 0.0193 \\
             Ours & \textbf{29.64} & \textbf{0.981} & \textbf{0.0183} \\
            \bottomrule
\end{tabular}
\end{small}
\label{Tab:ablas_generated_image}
\end{table}

\begin{figure}[ht]
\centering
    \includegraphics[width=1\columnwidth]{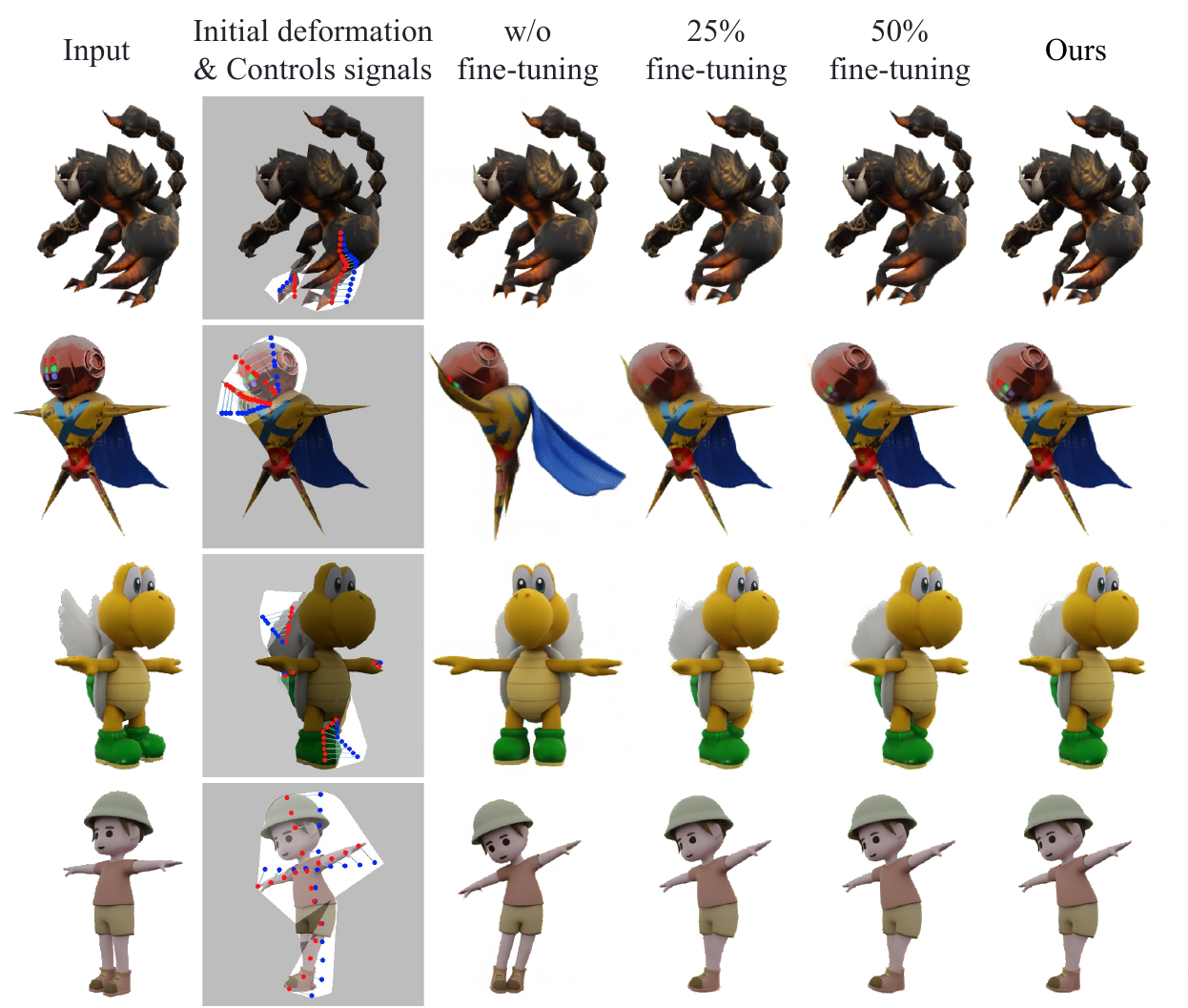}\vspace{-0.4cm} 
    \caption{\revise{Visualization of control signals and images generated with and without model fine-tuning.}} 
    \label{fig:drag_condition} 
\end{figure}

\vspace{0.5em}
\mypara{Effectiveness of Fine-tuning} 
To evaluate the impact of our fine-tuning strategy, we conducted an ablation study comparing the performance of the base model with and without the fine-tuning process. \revise{To further assess the effect of training data size on model fine-tuning, we evaluate image generation quality after fine-tuning on 25\% and 50\% of the training objects.} 
Fig.~\ref{fig:drag_condition} illustrates the control signals and corresponding images generated by the original model~\cite{xie2026TVEdit} \revise{without fine-tuning, variants fine-tuned with different amounts of training data, and our full model.}
Furthermore, Table~\ref{Tab:ablas_generated_image} provides a quantitative evaluation of the generated images by comparing them against the ground-truth images.
\revise{As shown, fine-tuning with only 25\% of the training data already substantially improves the alignment between the generated images and the target skeletons, while our full model achieves the best performance.}

\vspace{0.5em}
\mypara{\revise{Efficiency and Effectiveness Analysis of Skinning Volume}} \revise{We evaluate the effects of skinning volume resolution on deformation quality and optimization speed, and compare our sparse optimization strategy with a dense volume implementation. As shown in Table~\ref{Tab:ablas_skinning_volume}, sparse optimization substantially accelerates optimization at high resolutions, while our default resolution of \(256^3\) offers a good trade-off between accuracy and speed.
}

\begin{table*}[t]
    \caption{\revise{
    Quantitative comparison under imperfect skeletons. }}\vspace{-0.25cm}
    \label{Tab:skeleton_error}
    \centering
    \setlength{\tabcolsep}{3.1pt}
    \footnotesize
    \begin{tabular}{l cccc cccc cccc}
        \toprule
        \multirow{2}{*}{Method}
        & \multicolumn{4}{c}{Mild}
        & \multicolumn{4}{c}{Moderate}
        & \multicolumn{4}{c}{Severe} \\
        \cmidrule(lr){2-5}
        \cmidrule(lr){6-9}
        \cmidrule(lr){10-13}
        & PSNR$\uparrow$ & SSIM$\uparrow$
        & LPIPS$\downarrow$ & MSE$\downarrow$
        & PSNR$\uparrow$ & SSIM$\uparrow$
        & LPIPS$\downarrow$ & MSE$\downarrow$
        & PSNR$\uparrow$ & SSIM$\uparrow$
        & LPIPS$\downarrow$ & MSE$\downarrow$ \\
        \midrule

        \multicolumn{13}{l}{\textit{Joint Displacement}} \\
        BBW$^\dagger$
        & 22.12 & 0.930 & 0.0605 & 6.542
        & 21.49 & 0.922 & 0.0654 & 7.320
        & 19.33 & 0.897 & 0.0840 & 12.964 \\
        UniRig
        & 23.82 & 0.944 & 0.0536 & 3.188
        & 22.33 & 0.930 & 0.0611 & 3.850
        & 19.54 & 0.899 & 0.0846 & 12.835 \\
        Puppeteer
        & 24.23 & 0.948 & 0.0525 & 3.131
        & 22.54 & 0.932 & 0.0605 & 3.948
        & \textbf{19.83} & 0.902 & 0.0830 & 9.551 \\
        Ours
        & \textbf{24.38} & \textbf{0.950}
        & \textbf{0.0485} & \textbf{2.232}
        & \textbf{22.67} & \textbf{0.935}
        & \textbf{0.0560} & \textbf{2.931}
        & 19.82 & \textbf{0.907}
        & \textbf{0.0768} & \textbf{9.289} \\
        \midrule
        \multicolumn{13}{l}{\textit{Missing Joint Branches}} \\
BBW$^{\dagger}$
& 22.19 & 0.935 & 0.0581 & 5.600
& 20.98 & 0.927 & 0.0651 & 12.155
& 20.32 & 0.923 & 0.0687 & 18.908 \\
UniRig
& 24.54 & 0.953 & 0.0518 & 5.283
& 22.76 & 0.944 & 0.0597 & \textbf{11.629}
& 21.88 & 0.938 & 0.0647 & \textbf{16.782} \\
Puppeteer
& 25.18 & 0.958 & 0.0507 & 5.444
& 23.08 & 0.947 & 0.0598 & 14.047
& 22.07 & 0.942 & 0.0639 & 20.537 \\
Ours
& \textbf{26.10} & \textbf{0.964}
& \textbf{0.0445} & \textbf{2.226}
& \textbf{23.66} & \textbf{0.953}
& \textbf{0.0532} & 12.282
& \textbf{22.47} & \textbf{0.946}
& \textbf{0.0586} & 19.132 \\
\midrule

\multicolumn{13}{l}{\textit{Redundant Leaf Joints}} \\
BBW$^{\dagger}$
& -- & -- & -- & --
& -- & -- & -- & --
& -- & -- & -- & -- \\
UniRig
& 25.40 & 0.958 & 0.0486 & 2.688
& 25.22 & 0.956 & 0.0494 & 3.184
& 25.12 & 0.956 & 0.0497 & 3.151 \\
Puppeteer
& 26.05 & 0.962 & 0.0475 & 3.020
& 25.95 & 0.961 & 0.0477 & 3.064
& 25.82 & 0.960 & 0.0482 & 2.637 \\
Ours
& \textbf{26.60} & \textbf{0.966}
& \textbf{0.0431} & \textbf{1.948}
& \textbf{26.37} & \textbf{0.964}
& \textbf{0.0439} & \textbf{2.188}
& \textbf{26.19} & \textbf{0.964}
& \textbf{0.0446} & \textbf{2.231} \\
        \bottomrule
    \end{tabular}%

    \vspace{2pt}
    \begin{minipage}{\textwidth}
        \footnotesize
        $^\dagger$BBW succeeds on 24/22/24 cases for joint
displacement, 32/32/32 for missing joint branches, and
0/0/0 for redundant leaf joints, respectively, under
mild/moderate/severe perturbations (out of 40 cases
per condition). Its averages therefore use different
subsets and are not directly comparable to the
full-set averages of the other methods.
``--'' indicates no successful cases. ``$\uparrow$" (resp. $\downarrow$) indicates the larger (resp. smaller), the better. MSE values are reported in units of $10^{-4}$. The best results are highlighted in bold.
    \end{minipage}
\end{table*}

\begin{table}[h]
\caption{\revise{Effect of skinning volume resolution on skinning accuracy, runtime, and peak GPU memory for dense and sparse implementations on Articulation-XL 2.0.}}
\vspace{-0.25cm}
\label{Tab:ablas_skinning_volume}
\centering
\resizebox{\linewidth}{!}{
\setlength{\tabcolsep}{3.1pt}
\begin{small}
\begin{tabular}{l c c c c c c c c c}
\toprule
Method & PSNR$^{\uparrow}$ & SSIM$^{\uparrow}$
& LPIPS$^{\downarrow}$ & MSE$^{\downarrow}$
& $D_{\text{Max}}^{\downarrow}$ & CD$^{\downarrow}$
& L1$^{\downarrow}$ & Time (s) & {Mem.} \\
\midrule
\textit{i) dense} \\
64  & 26.53 & 0.966 & 0.0428 & 1.95 & 0.14 & 4.06 & 0.51 & 94.62  & {2.99} \\
128 & 26.57 & 0.966 & 0.0426 & 1.93 & 0.14 & 4.04 & 0.51 & 151.51 & {3.67} \\
256 & 26.56 & 0.966 & 0.0427 & 1.94 & 0.14 & 4.03 & 0.51 & 481.36 & {18.88} \\
\midrule
\textit{ii) sparse} \\
64  & 26.54 & 0.966 & 0.0428 & 1.94 & 0.14 & 4.05 & 0.51 & 91.88  & {2.97} \\
128 & 26.55 & 0.966 & 0.0426 & 1.93 & 0.14 & 4.04 & 0.51 & 96.58  & {3.11} \\
256 & 26.58 & 0.966 & 0.0426 & 1.93 & 0.14 & 4.02 & 0.51 & 103.96 & {3.26} \\
512 & 26.60 & 0.966 & 0.0426 & 1.94 & 0.14 & 4.00 & 0.50 & 111.80 & {3.48} \\
\bottomrule
\end{tabular}
\end{small}
}

\begin{minipage}{\linewidth}
\vspace{2pt}
\footnotesize
MSE and CD are reported in units of $10^{-4}$ and $10^{-3}$,
respectively. Time denotes the average total runtime for
initialization and motion optimization.
{Mem. (GiB) denotes the maximum peak GPU memory allocated across all scenes during the optimization stage.}
$\uparrow$ and $\downarrow$ indicate that higher and lower
values are better, respectively.
\end{minipage}
\end{table}

\noindent\textbf{Dependency Analysis of Skeleton.}
\revise{We evaluate robustness to imperfect skeletons on 40 Articulation XL 2.0
objects under three perturbations: joint displacement, missing branches,
and redundant joints, each at mild, moderate, and severe levels.
We displace joints by 1\%, 2\%, or 5\% of the rest-mesh bounding-box
diagonal, remove progressively larger anatomical subtrees, or append
increasing numbers of terminal bones with varying lengths.
As shown in Table~\ref{Tab:skeleton_error}, our method achieves better performance under these skeleton perturbations, demonstrating greater robustness to imperfect skeletons. Detailed construction of imperfect skeletons and qualitative results are provided in Appendix~\ref{app:skeletons}.}

\vspace{0.5em}
\mypara{\revise{Runtime Analysis}}
\revise{For the runtime, generating the guidance images takes approximately 250 seconds (2.5 seconds per image). For the mesh-anchored configuration, skinning optimization takes 104 seconds on average, including 27 seconds for skinning initialization. Among the baselines, BBW, UniRig, and Puppeteer take 12.42, 8.67, and 15.62 seconds per asset, respectively.  Our method takes longer than the baselines because it requires multi-view image generation followed by 3D optimization guided by these images. Nevertheless, the total runtime is still acceptable for an offline skinning task, where high accuracy is a priority. In Appendix~\ref{app:runtime}, we further analyze how optimization and inference times vary with the number of Gaussians.
}

\section{Conclusion and Discussion}
\label{sec:conclusion}

In this paper, we presented G-Skin, a novel framework that enables high-quality animation of arbitrary 3D Gaussian primitives by automatically learning skinning weights for a given skeleton. 
A key innovation of our work is the generative skinning pipeline, which circumvents the lack of specialized datasets by leveraging large-scale vision foundation models to provide skeleton-controllable guidance. Through a stable optimization strategy and tailored geometric regularizations, G-Skin demonstrates exceptional generalization capabilities across a wide range of arbitrary, complex objects without the need for predefined templates. 
Furthermore, our method can be flexibly applied to standard 3D Gaussian representation as well as augmented variants, such as mesh-anchored 3D Gaussian representation.
Experimental results validated that G-Skin achieves superior animation performance compared to existing approaches, providing a scalable and efficient solution for the next generation of animatable 3D content.

\vspace{0.5em}
\noindent \textbf{Limitations and Future Work.}
Despite its effectiveness, G-Skin has certain limitations. 
\revise{First, geometric sticking between kinematically distinct regions (e.g., between the body and associated props, as shown in Fig.~\ref{fig:failure_case}) remains challenging. While our method reduces deformation artifacts compared with existing methods, some local artifacts may persist. This limitation arises from our geometric regularization within small spatial neighborhoods, which may inadvertently couple the motion of geometrically adjacent but kinematically distinct parts.}
Second, owing to the reliance on 2D generative priors, our framework requires test-time optimization to propagate 2D knowledge into the 3D space, which precludes direct, end-to-end prediction of 3D skinning weights.
Finally, our current formulation does not explicitly optimize 3D Gaussian parameters to reduce rendering artifacts during deformation, such as in unobserved regions like the armpits.

\begin{figure}[t]
   \centering
    \includegraphics[width=0.95\columnwidth]{figs/failure_case.pdf}\vspace{-0.4cm} 
    \caption{\revise{Example of failure caused by improper sticking between geometric surfaces.}} \vspace{-0.3cm}
    \label{fig:failure_case}
\end{figure}

Future work will focus on exploring optimized 3D Gaussian distributions or novel continuous representations that mitigate rendering artifacts during animation. Additionally, we aim to leverage 2D generative priors to synthesize high-quality 3D datasets, thereby facilitating the development of a fully end-to-end, feed-forward rigging framework of 3D Gaussian representation.

\bibliographystyle{IEEEtran}
\bibliography{egbib}

\begin{IEEEbiography}[{\includegraphics[width=1in]{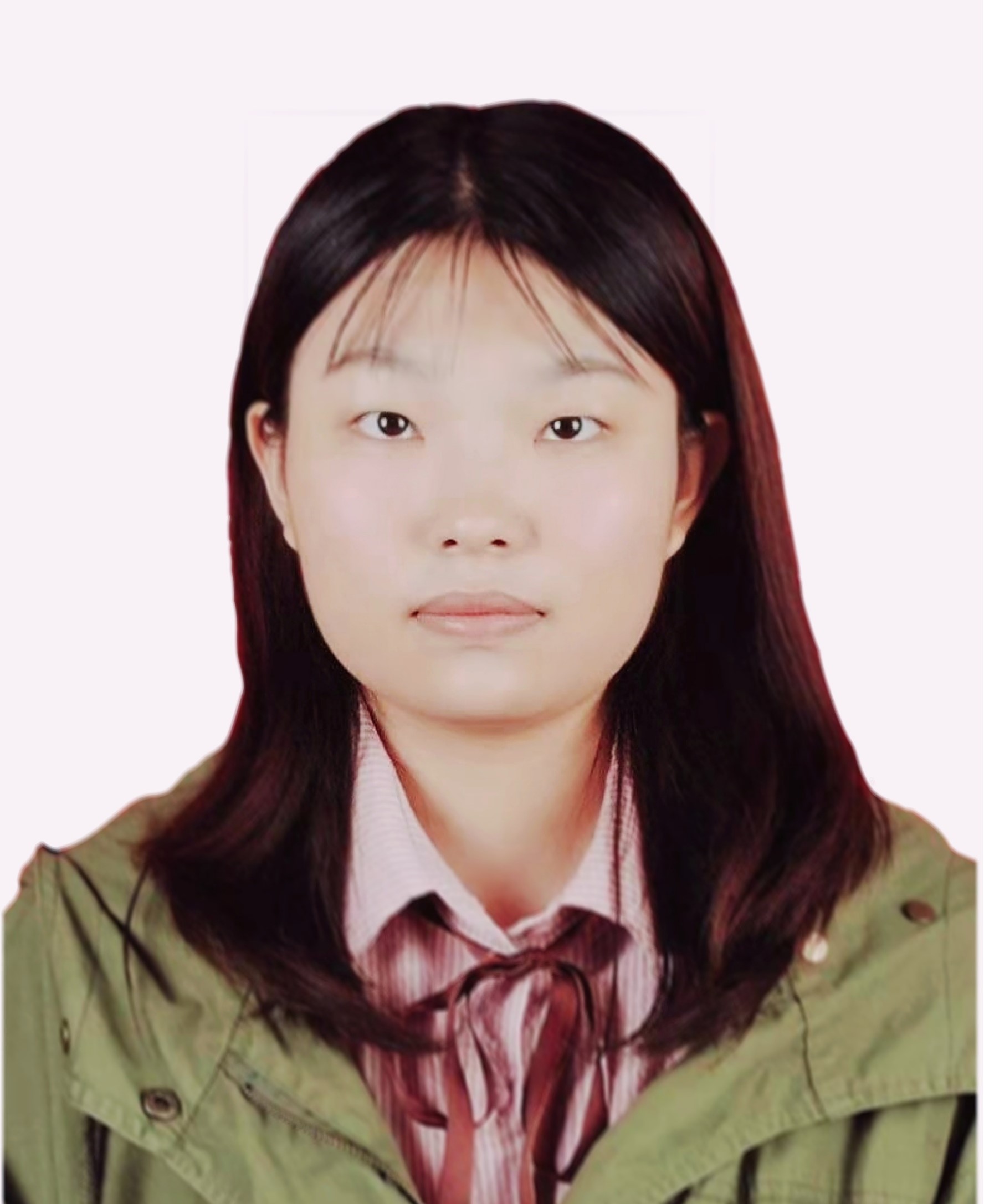}}]{Yuxin Yao} is currently a postdoctoral researcher with the Department of Computer Science, City University of Hong Kong. She received the BE degree from the University of Electronic Science and Technology of China, Chengdu,
China, in 2018, and the PhD degree from the University of Science and Technology of China, Hefei, China, in 2023. Her research interests include 3D vision and computer graphics.
\end{IEEEbiography}

\begin{IEEEbiography}[{\includegraphics[width=1in]{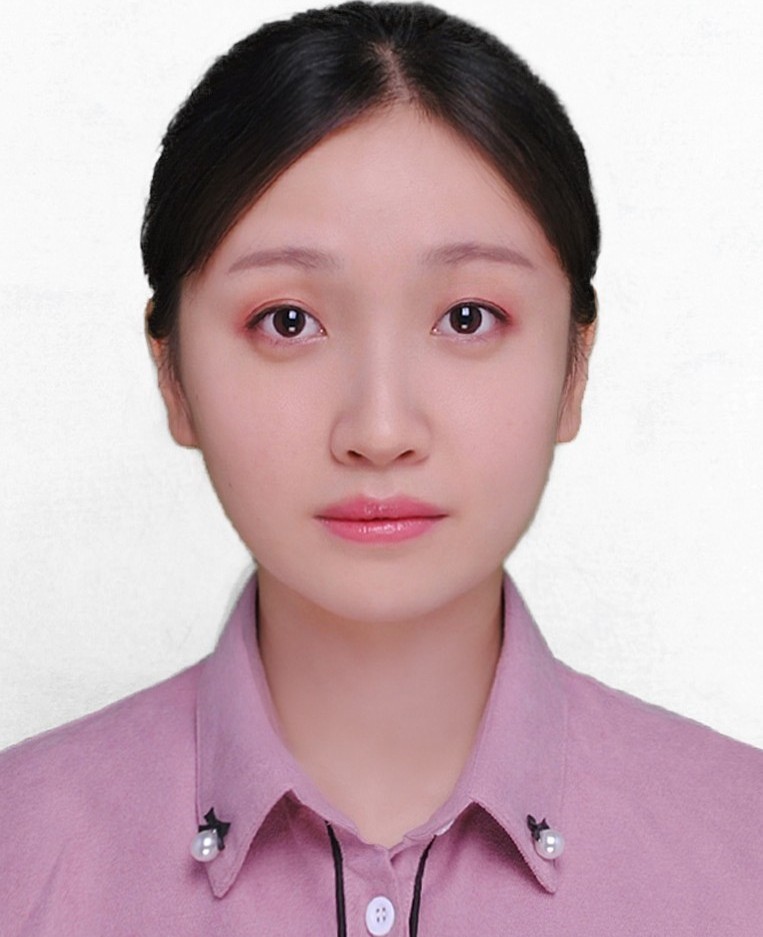}}]{Kendong Liu} is currently a PhD student with the Department of Computer Science, City University of Hong Kong. She received the BE degree from Xidian University, Xi’an, China, in 2020, and the MS degree from Xidian University, Xi’an, China, in 2023. Her research interests include generative models and 3D vision.
\end{IEEEbiography}

\begin{IEEEbiography}[{\includegraphics[width=1in]{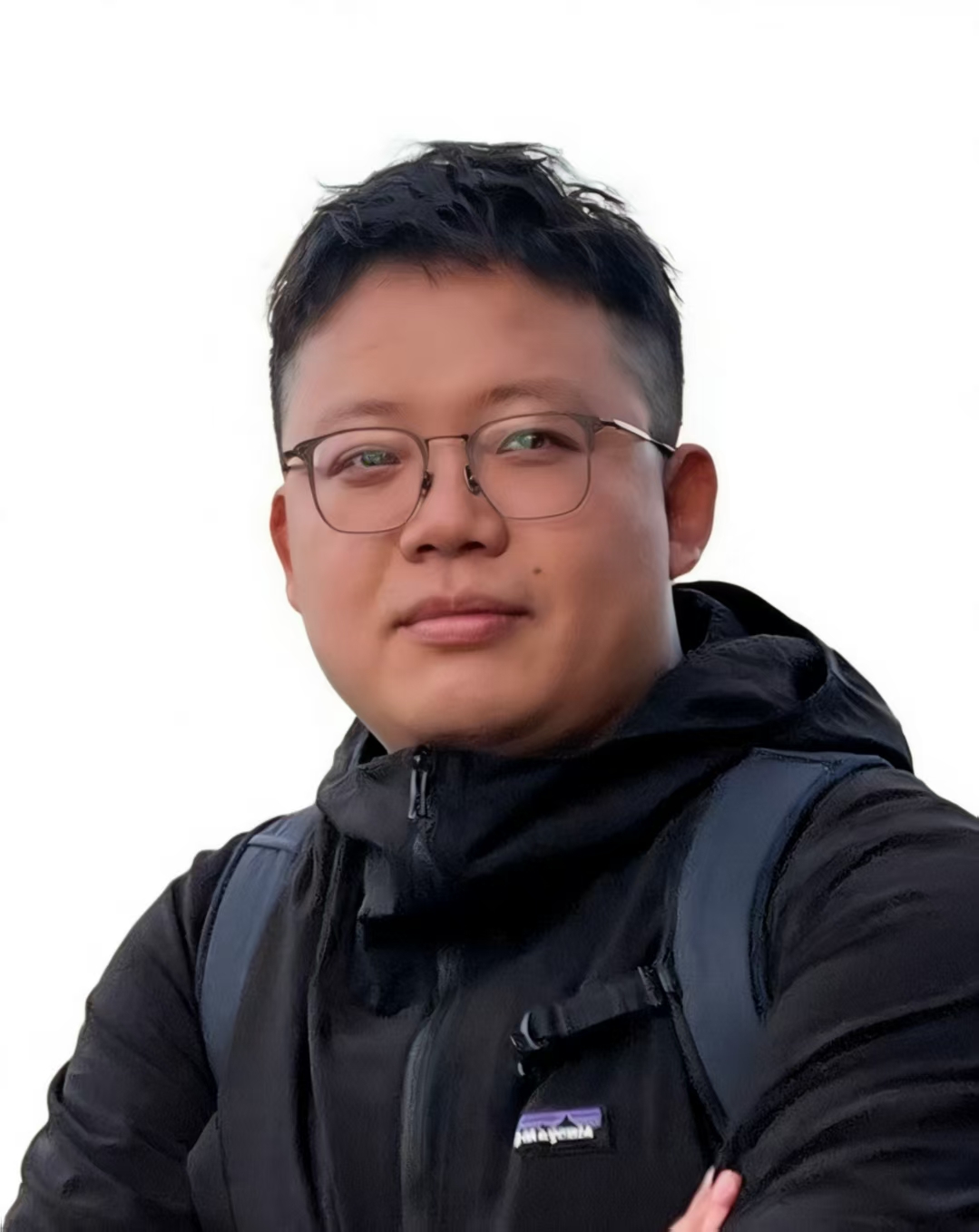}}]{Shiqi Zhou} is the Head of the Physical Simulation
\& Animation Team at the Central Media Technology
Institute, Huawei. He holds a Ph.D. in Computational Mathematics from the State Key Laboratory
of Scientific and Engineering Computing (LSEC),
Chinese Academy of Sciences. His research focuses
on AI+CG \& 3DV, including 3D/4D generation and
editing, physical simulation. 
\end{IEEEbiography}

\begin{IEEEbiography}[{\includegraphics[width=1in]{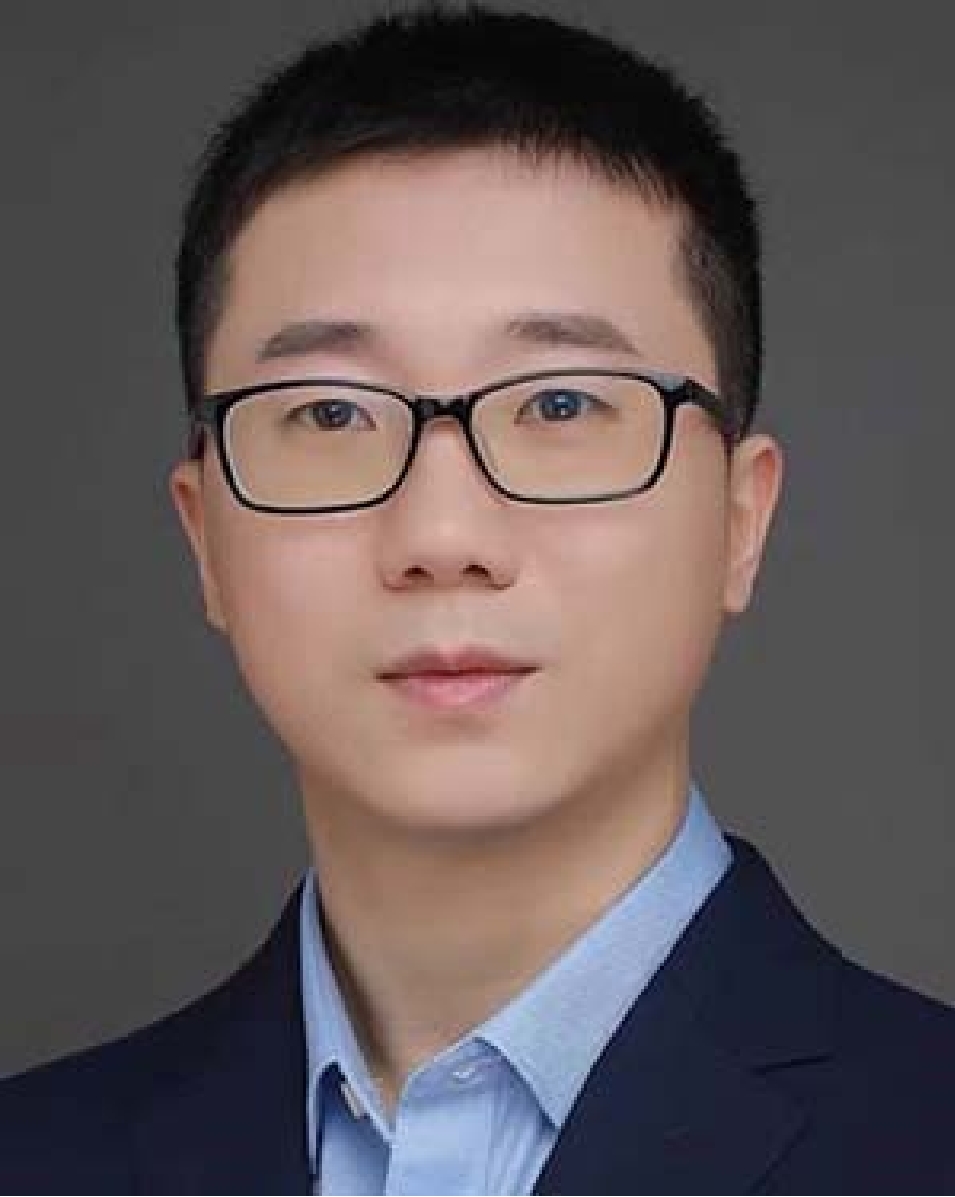}}]{Jiazhi Xia} (Member, IEEE) is currently a professor with the School of Computer Science and Engineering, Central South University. His research interests include visualization, geometry processing, and 3D vision. 
\end{IEEEbiography}

\begin{IEEEbiography}[{\includegraphics[width=0.9in]{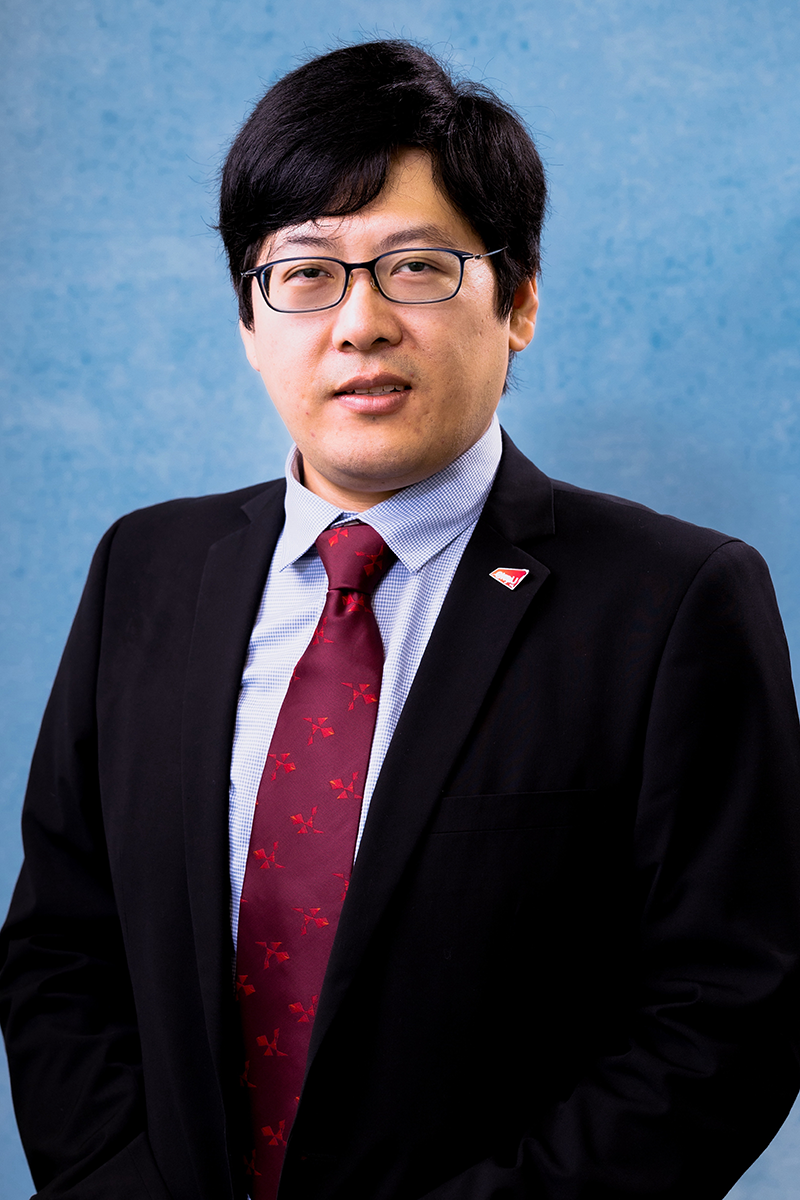}}]{Junhui Hou} is a Professor with the Department of Computer Science, City University of Hong Kong (CityUHK). His research interests include multidimensional visual computing and machine learning.

He received the Early Career Award and the Research Fellow Award from the Hong Kong Research Grants Council, the Excellent Young Scientists Fund from the National Natural Science Foundation of China, the IEEE TIP Best Paper Award, and the CityUHK Presidential Research Excellence Award for Junior Faculty. He serves as a Senior Area Editor for IEEE TIP and as an Associate Editor for IEEE TVCG and TMM. He served as an Associate Editor for IEEE TIP and TCSVT.
\end{IEEEbiography}

\clearpage
\appendices

\section{\revise{More analysis and discussion}}
\label{app:analysis}

\vspace{0.5em}
\subsection{\revise{Reliability of Small Articulated Parts}}
\label{app:small_parts}
\revise{To evaluate our method across articulated parts of different sizes, we group joints into small, medium, and large categories and report their deformation errors separately in Table~\ref{Tab:local_error_bone_level}. Joint size is defined by the surface-area fraction of its influence region, determined using the ground-truth skinning weights. Specifically, each vertex is assigned to the joint with the highest ground-truth weight, forming a region \(R_j\). We estimate each vertex’s area as one-third of the total area of its incident triangles and sum these contributions to obtain the region’s area. Regions occupying less than \(1\%\), \(1\%\)–\(5\%\) (excluding \(5\%\)), and at least \(5\%\) of the total mesh surface area are classified as small, medium, and large, respectively. To account for differences in part size, we compute the surface-area-weighted deformation RMSE against the ground-truth animated meshes over the test poses and normalize it by the diagonal length of each region’s canonical bounding box, reporting the result as a percentage. Errors are averaged over joints within each category for each asset, followed by averaging across assets. As shown in Table~\ref{Tab:local_error_bone_level}, smaller joints tend to exhibit larger relative errors. Nevertheless, apart from a minor difference in the medium-size category, our method consistently achieves lower errors than the baselines, demonstrating its reliability on small joints. This robustness benefits from our geometric coupling loss, which remains effective even when joints occupy only a small image area.}

\begin{table}[htb]
	\caption{\revise{Comparison of local deformation accuracy across articulated-part sizes.} 
    }\vspace{-0.25cm}
	\label{Tab:local_error_bone_level}
		\setlength{\tabcolsep}{3.1pt}
	\centering
	\begin{small}
		\begin{tabular}{ l  c  c  c }
            \toprule 
             Method  & small & medium & large  \\ %
            \midrule
         BBW~\cite{jacobson2011bounded} & 11.42 & 8.46 & 7.50 \\
            UniRig~\cite{zhang2025unirig} & 10.91 & 5.71 & 3.81 \\%
            Puppeteer~\cite{song2025puppeteer} & 10.78 & \textbf{4.39} & 3.04  \\
            Ours & \textbf{10.10} & 4.71 & \textbf{2.76} \\ 
            \bottomrule
		\end{tabular}
	\end{small}
    
\begin{minipage}{\linewidth}
\vspace{2pt}
\footnotesize
Results are reported as relative RMSE (\%). Lower values are better.
The best results are highlighted in bold.
\end{minipage}
\end{table}

\subsection{\revise{Construction of Imperfect Skeletons}}
\label{app:skeletons}
\revise{Here, we provide additional details on the construction of imperfect skeletons described in Sec. \textcolor{red}{IV-D} of the main paper.
Let $D_{\text{AABB}}$ denote the axis-aligned bounding-box diagonal of the original
rest-mesh vertices. The original geometry, ground-truth animations,
and evaluation cameras remain unchanged.}

\vspace{0.5em}
\noindent\revise{
\textbf{Joint displacement.}
We keep both endpoints of the root bone fixed and displace
each remaining joint along a random unit direction by
$0.01\cdot D_{\text{AABB}}$, $0.02\cdot D_{\text{AABB}}$, or $0.05\cdot D_{\text{AABB}}$ for mild, moderate, or severe
perturbations, respectively. Coincident endpoints move together.
For each joint, we use the same sampled direction across
severity levels and vary only the displacement magnitude.
The hierarchy and original local rotation channels are retained,
and forward kinematics for original motion sequences is recomputed using the perturbed skeleton. }

\vspace{0.5em}
\noindent\revise{
\textbf{Missing branches.}
We simulate missing joints by removing a selected bone and
all its descendants. For a typical limb, the mild setting
removes only the hand or foot, the moderate setting removes
the forearm or lower leg together with the hand or foot,
and the severe setting removes the entire arm or leg.
Each higher severity level includes all bones removed at
the preceding level and removes additional bones.
For eligible humanoid rigs, we select the affected limb
using a fixed random seed, with approximately equal numbers
of objects assigned to the left arm, right arm, left leg,
and right leg. For other rigs, we define the removal sets
according to their rest-pose structure and bone connectivity.
Two special cases accommodate simpler skeletons.
For limbs with only two segments, the severe setting also
removes the opposite limb. For unbranched chains, the three
levels remove the last one, two, or three bones, respectively.
All removal sets are determined before evaluating any method.
The global root is never removed, and the remaining bones
keep their original rest positions, parent-child relationships,
and local motion parameters.
}

\vspace{0.5em}
\noindent\revise{
\textbf{Redundant joints.}
We simulate redundant joints by adding terminal bones to the
original skeleton. Let $N_{\mathcal{B}}$ denote the original
number of bones. The total numbers of added bones at the mild,
moderate, and severe levels are
\[
n_1=\max(2,\lceil0.10\cdot N_{\mathcal{B}}\rceil)\]
\[
n_2=\max(n_1+1,\lceil0.25\cdot N_{\mathcal{B}}\rceil),\]
\[n_3=\max(n_2+1,\lceil0.50\cdot N_{\mathcal{B}}\rceil).
\]
These counts correspond approximately to 10\%, 25\%, and 50\%
of the original bone count, while ensuring at least two additions
at the mild level and a strict increase at each subsequent level.
Each added bone is attached to a distinct parent sampled uniformly
from the original bones, including internal bones and the root.
It starts at its parent's tail and extends in an independently
sampled random direction.
Higher severity levels retain all previously added bones unchanged
and introduce additional bones. The lengths of these new additions
are distributed across
$[0.03\cdot D_{\text{AABB}},0.10\cdot D_{\text{AABB}}]$,
$[0.03\cdot D_{\text{AABB}},0.20\cdot D_{\text{AABB}}]$, and
$[0.03\cdot D_{\text{AABB}},0.35\cdot D_{\text{AABB}}]$
at the mild, moderate, and severe levels, respectively, using
stratified sampling. 
The original motion parameters remain unchanged. 
Added bones have identity local rotations in every pose, so they move
with their parents without any additional local rotation.
}

\begin{figure*}[ht] 
    \centering
    \includegraphics[width=1.0\textwidth]{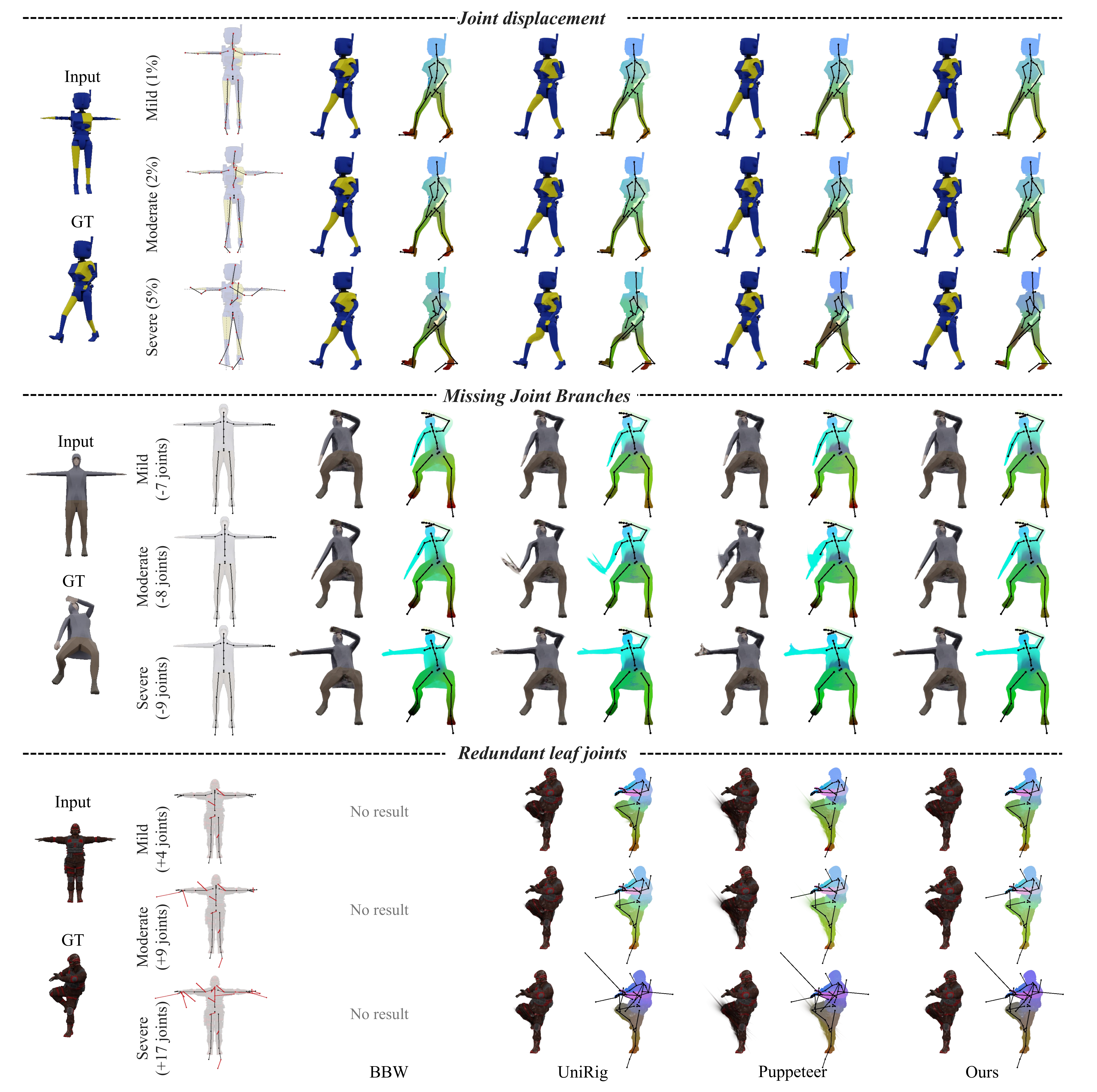}\vspace{-0.4cm}  
    \caption{\revise{Performance of our method under imperfect skeletons.}
    }\label{fig:failure_skeleton} 
\end{figure*}

\revise{Fig.~\ref{fig:failure_skeleton} illustrates examples of imperfect skeletons with different perturbation types and severity levels, along with the corresponding results of our method and the baselines.}

\begin{figure}[ht] 
    \centering
    \includegraphics[width=1\columnwidth]{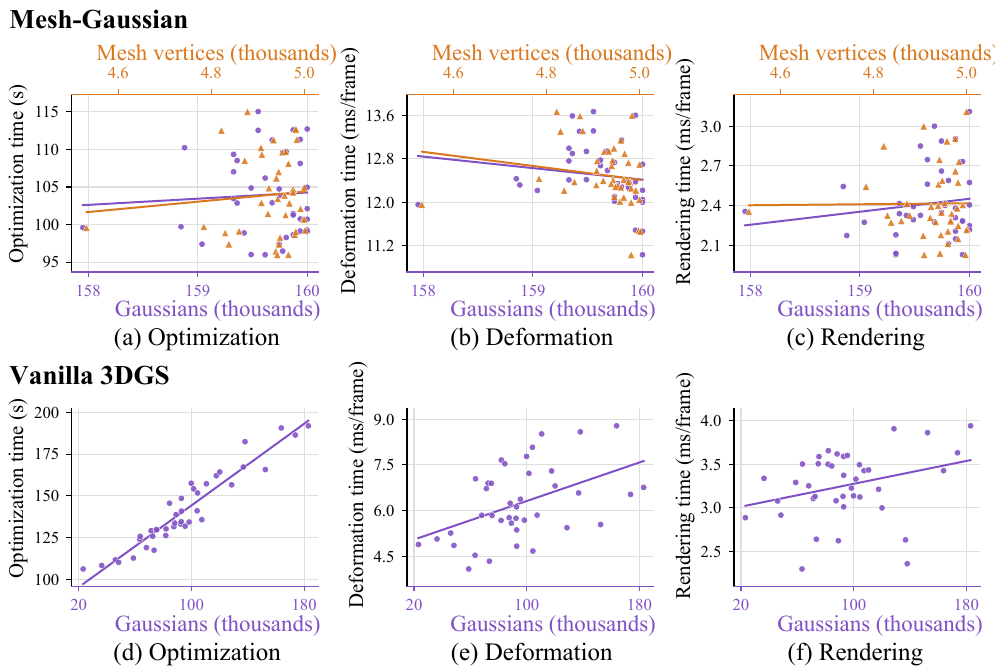} 
    \caption{\revise{Runtime analysis for mesh-based Gaussians (top) and vanilla 3DGS (bottom): skinning optimization (a, d), per-frame deformation (b, e), and rendering (c, f) during inference. Purple and orange markers denote Gaussian and mesh-vertex counts, respectively. Each marker represents one object; solid lines indicate linear fits.}
    }\label{fig:runtime_analysis} 
    \vspace{-0.4cm} 
\end{figure}

\revise{
\subsection{Runtime Analysis with Respect to Gaussian Count}
\label{app:runtime}
We further analyze skinning optimization time and per-frame deformation and rendering time on the Articulation-XL 2.0 test set, as shown in Fig.~\ref{fig:runtime_analysis}. For pure 3DGS, optimization time increases approximately linearly with the Gaussian count, while inference times exhibit weaker positive trends with greater variation across objects. For mesh-based Gaussians, deformation is computed through a low-resolution mesh proxy and transferred to the Gaussians, making runtime dependent on both mesh and Gaussian complexity. Since both counts span relatively narrow ranges in this test set, we observe no clear increase in optimization or inference time with Gaussian count. On average, deformation and rendering take 12.50 and 2.41 ms per frame for mesh-based Gaussians, and 6.23 and 3.26 ms per frame for pure 3DGS, respectively.
}

\begin{figure}[ht] 
    \centering
    \includegraphics[width=\columnwidth]{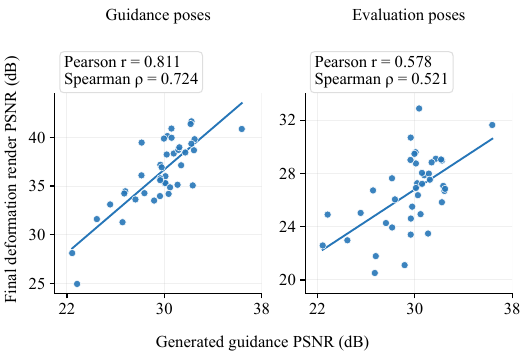} \vspace{-0.4cm} 
    \caption{
    \revise{Relationship between generated guidance quality and
skinning-driven rendering accuracy at guidance poses (left)
and evaluation poses (right). Each point represents one object;
solid lines indicate linear fits. Higher guidance-image PSNR
correlates with higher rendering PSNR in both settings,
suggesting that better guidance benefits deformation accuracy
beyond the guidance poses.}}
    \label{fig:guidance_quality} 
\end{figure}

\revise{
\subsection{Effect of Guidance Quality on Skinning Accuracy}
\label{app:guidance_quality}
To examine how the quality of generated guidance relates to
skinning accuracy, we compare guidance-image PSNR with the
rendering PSNR of the learned skinning under the corresponding
guidance poses and the prescribed evaluation poses.
At the guidance poses, the reference renderings are obtained by
directly applying the ground-truth skinning weights at the same
poses. At the evaluation poses, the reference images are frames
from ground-truth mesh animation videos rendered in Blender.
As shown in Fig.~\ref{fig:guidance_quality}, higher guidance
quality is associated with higher rendering accuracy in both
settings. The correlation is stronger for the guidance poses
(Pearson $r=0.811$, Spearman $\rho=0.724$) and remains positive
for the evaluation poses ($r=0.578$, $\rho=0.521$).
Pose-dependent lighting and shading in the Blender-rendered
reference images may introduce additional photometric discrepancies
beyond skinning errors, which may partly explain the weaker
correlation at the evaluation poses.
These results suggest that improvements in image-space guidance
can translate into more accurate skinning-driven deformations,
with benefits extending beyond fitting the guidance images.
This analysis supports the importance of accurate generated guidance for the final deformation quality.
}

\begin{figure*}[ht] 
    \centering
    \includegraphics[width=\textwidth]{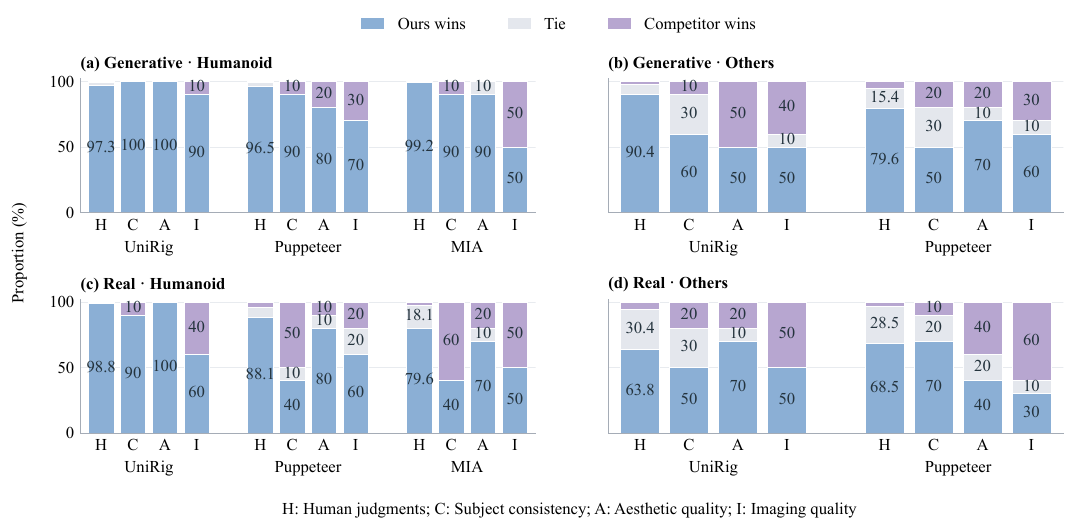} \vspace{-0.4cm} 
    \caption{
    \revise{Comparison of user-study results (H) and VBench metrics: subject consistency (C), aesthetic quality (A), and imaging quality (I). Stacked bars show win, tie, and loss proportions for Ours against each competitor.}}
    \label{fig:vbench_userstudy} 
\end{figure*}

\revise{
\subsection{Reliability of VBench Evaluation}
\label{app:vbench}
VBench measures animation quality indirectly rather than skinning correctness. To assess its agreement with human preferences, we compared the win/tie/loss distributions from its metrics with those from our pairwise user study.}

\revise{
Specifically, the study includes 40 selected sequences: 10 humanoid and 10 non-humanoid sequences from each of the generative and real datasets. Comparing Ours against three competitors for humanoid assets and two for non-humanoid assets yields 100 paired comparisons. We collected 26 completed questionnaires, giving 2,600 judgments. Participants assessed shape-skeleton compatibility and deformation plausibility. For each comparison, we calculated the difference between the VBench scores of Ours and the competitor. Using scores normalized to \([0,1]\), an absolute difference below 0.001 was treated as a tie; otherwise, the higher-scoring method was considered the winner. Human win rates were computed as \(W/(W+T+L)\), retaining ties in the denominator. We report all three outcomes separately for each data category and competitor. 
}

\revise{
Figure~\ref{fig:vbench_userstudy} compares the user-study results
with three VBench metrics: subject consistency, aesthetic quality,
and imaging quality (IQ). The results show partial agreement,
with substantial variation across metrics and categories.
For generative humanoid assets, the human win rate for Ours
is 97.7\%, compared with 93.3\% for subject consistency and
90.0\% for aesthetic quality. For real humanoid assets,
aesthetic quality yields a win rate of 83.3\%, close to the
human rate of 88.8\%, whereas the other two metrics show
larger discrepancies. For real non-humanoid assets,
subject consistency yields 60.0\%, compared with 66.2\%
from human judgments. Agreement is weaker for generative
non-humanoid assets, where the human win rate is 85.0\%
but the VBench-based rates range from 55.0\% to 60.0\%.
We also observe that IQ exhibits substantial discrepancies
from human judgments across all four categories, with
absolute win-rate gaps of approximately 26--32 percentage
points. We therefore exclude IQ from our main quantitative
evaluation and retain subject consistency and aesthetic
quality as complementary metrics, while reporting all three here.
}

\revise{
These findings support using VBench as a complementary perceptual evaluation, while also demonstrating its limitations. 
For unseen generated assets, ground-truth skinning weights and deformed geometry are unavailable, preventing direct GT-based error evaluation. We therefore combine automated VBench measurements with human assessments of deformation plausibility, rather than treating VBench alone as evidence of correct skinning. 
}

\revise{Table~\ref{tab:vbench_winrates_agreement} additionally reports
VBench win/tie/loss rates across all evaluated generated 3D
assets and real-data examples. Each entry also includes
Human Agreement (HA), the percentage of individual human
judgments that agree with the corresponding VBench
win/tie/loss decisions on the user-study subset.
For each data category, competitor, and metric, HA is computed as
\[
\mathrm{HA}
=
\frac{\sum_{i=1}^{M}\sum_{j=1}^{P}
\mathbb{I}\!\left[\hat{y}_i=y_{ij}\right]}
{MP}\times100\%,
\]
where $M=10$ is the number of paired comparisons in the
corresponding user-study subset, $P=26$ is the number of
participants, $\hat{y}_i$ is the VBench result for
$i$-th comparison case, and $y_{ij}$ is participant $j$'s judgment.
Higher HA indicates closer alignment with human preferences.
For comparisons with high HA, Ours generally achieves
substantially more wins than losses. On generated humanoid
assets, HA ranges from 79.2\% to 97.3\%, and Ours wins
69.8--75.9\% of comparisons in subject consistency and
78.7--86.4\% in aesthetic quality against the three competitors.
On real humanoid assets, Ours outperforms UniRig in 95.8\%
and 91.7\% of comparisons for these two metrics, with HA
of 88.8\% and 98.8\%, respectively. These results further demonstrate the advantages of our method.}

\begin{table}[t]
\centering
\caption{\revise{VBench-based pairwise comparison results and agreement with human judgments.}}
\label{tab:vbench_winrates_agreement}
\setlength{\tabcolsep}{3pt}
\renewcommand{\arraystretch}{1.05}
\resizebox{\columnwidth}{!}{%
\begin{tabular}{lcc}
\toprule
Ours vs. &  Consis. & Aes. \\
\midrule
\multicolumn{3}{l}{\textit{Generated / Human}} \\
UniRig    & 73.3 / 7.7 / 19.0 (97.3)  & 86.4 / 2.6 / 11.0 (97.3) \\
Puppeteer &  75.9 / 11.4 / 12.7 (86.9) & 86.2 / 4.1 / 9.7 (79.2)  \\
MIA       &  69.8 / 13.9 / 16.3 (89.2) & 78.7 / 5.4 / 15.9 (89.2) \\
\midrule
\multicolumn{3}{l}{\textit{Generated / Non-human}} \\
UniRig    &  71.4 / 18.6 / 10.0 (65.0) & 61.4 / 8.6 / 30.0 (50.8)  \\
Puppeteer &  58.6 / 25.7 / 15.7 (51.2) & 61.4 / 17.1 / 21.4 (59.2) \\
\midrule
\multicolumn{3}{l}{\textit{Real / Human}} \\
UniRig    &  95.8 / 0.0 / 4.2 (88.8)   & 91.7 / 0.0 / 8.3 (98.8)  \\
Puppeteer &  50.0 / 12.5 / 37.5 (40.8) & 83.3 / 8.3 / 8.3 (75.4)  \\
MIA       &  45.8 / 8.3 / 45.8 (37.7)  & 54.2 / 12.5 / 33.3 (56.5) \\
\midrule
\multicolumn{3}{l}{\textit{Real / Non-human}} \\
UniRig    &  54.2 / 20.8 / 25.0 (46.2) & 66.7 / 8.3 / 25.0 (53.8)  \\
Puppeteer &  70.8 / 16.7 / 12.5 (64.2) & 50.0 / 20.8 / 29.2 (28.8) \\
\bottomrule
\end{tabular}%
}

\begin{minipage}{\linewidth}
\vspace{2pt}
\footnotesize
Results (\%) are reported as Win / Tie / Loss (HA) for Ours.
HA denotes agreement with human judgments on the corresponding
user-study subset. 
\end{minipage}
\end{table}

\revise{
\section{Visualization results}
\label{app:visualization}
In this section, we provide additional visual comparisons on generated 3D assets, including humanoid and non-humanoid characters in Figs.~\ref{fig:compare_generate_more} and \ref{fig:compare_generate_more_animal}, respectively. These results further demonstrate the superiority of our method. We also encourage readers to visit our project page \url{https://yaoyx689.github.io/gskin_materials/user_study.html} for additional video demonstrations regarding the user study. 
}

\begin{figure*}[ht] 
    \centering
    \includegraphics[width=1.0\textwidth]{figs_supp/compare_gene_more_h.pdf}
    \vspace{-0.5cm}
    \caption{
    \revise{Visual comparisons with the state-of-the-art methods on the generative humanoid 3D assets.} }
    \label{fig:compare_generate_more} 
\end{figure*}

\begin{figure*}[ht] 
    \centering
    \includegraphics[width=1.0\textwidth]{figs_supp/compare_gene_more_a.pdf} \vspace{-0.4cm} 
    \caption{
     \revise{Visual comparison with the state-of-the-art methods on the generative non-humanoid 3D assets.}}
    \label{fig:compare_generate_more_animal} 
\end{figure*}

\begin{figure}[ht] 
    \centering
    \includegraphics[width=0.8\columnwidth]{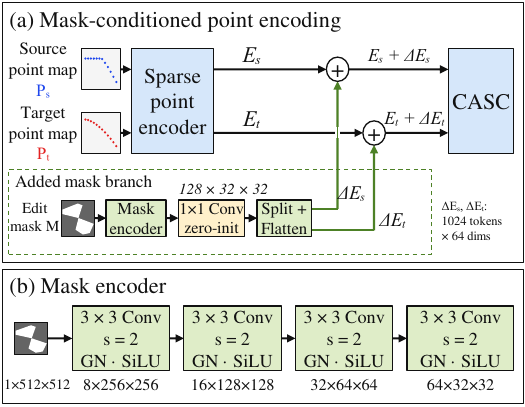} \vspace{-0.3cm} 
    \caption{
    \revise{Network architecture of the mask-conditioned point encoding module.}}
    \label{fig:network_mask_encoding} 
\end{figure}

\revise{
\section{Details of Skeleton-Controllable Guidance Generation Model}
\label{app:guidance_model}
\mypara{Mask-Conditioned Point Encoder}
We augment the original TVEdit~\cite{xie2026TVEdit} sparse point encoder with a lightweight, learnable mask-conditioning branch to provide explicit editing-region guidance. The input is a binary mask \(M\in\{0,1\}^{H\times W}\), spatially aligned with the source image, where 1 denotes regions permitted to change and 0 denotes regions intended to remain unchanged. 
The branch comprises four downsampling blocks, each consisting of a \(3\times3\) convolution with stride 2, GroupNorm, and SiLU, with output channels of 8, 16, 32, and 64, respectively. A subsequent \(1\times1\) convolution projects the features to 128 channels, which are split into two 64-channel feature maps. These maps are flattened into spatial tokens and added as residuals to the source-point and target-point embeddings before being passed to the original Content-Aware Spatial Controller (CASC). For a \(512\times512\) input, each residual comprises \(32\times32\) tokens. The final projection is zero-initialized, preserving the original point embeddings at initialization. The branch introduces 32,824 trainable parameters and is trained using the existing flow-velocity prediction mean squared error objective, without an additional mask-specific loss. The mask serves as a learned conditioning signal, and no hard attention gating or output compositing is applied. Fig.~\ref{fig:network_mask_encoding} illustrates the architecture of the mask-conditioned point encoder.
}

\revise{We compare our mask-conditioned model with the original TVEdit architecture fine-tuned without mask conditioning. Both models are evaluated on the same 4,000 guidance images from the 40 Objaverse assets, using identical source images, control points, and inference settings. As shown in Table~\ref{tab:mask_guidance_quality}, our model achieves higher PSNR and SSIM and lower LPIPS, demonstrating improved accuracy and perceptual similarity of the generated guidance images to the ground truth.}

\begin{table}[t]
\centering
\caption{\revise{Comparison of guidance image quality. The best results are highlighted in bold.}}
\label{tab:mask_guidance_quality}
\begin{tabular}{l c c c}
\toprule
Method & PSNR$\uparrow$ & SSIM$\uparrow$ & LPIPS$\downarrow$ \\
\midrule
w/o mask & 29.26 & 0.978 & 0.0204 \\
Ours & \textbf{29.64} & \textbf{0.981} & \textbf{0.0183} \\
\bottomrule
\end{tabular}
\end{table}

\vfill

\end{document}